\documentclass{article} 
\usepackage{iclr2026_conference,times}

\usepackage{amsmath,amsfonts,bm}

\def\eqref#1{equation~\ref{#1}}

\def\1{\bm{1}}

\DeclareMathAlphabet{\mathsfit}{\encodingdefault}{\sfdefault}{m}{sl}
\SetMathAlphabet{\mathsfit}{bold}{\encodingdefault}{\sfdefault}{bx}{n}

\usepackage{microtype}
\usepackage{hyperref}
\usepackage{url}
\usepackage{booktabs}

\usepackage{lineno}
\usepackage{bbm}
\usepackage{graphicx}
\usepackage{subcaption}
\usepackage{amsmath}
\usepackage{enumitem}
\usepackage{cleveref}
\usepackage{setspace}
\usepackage{colortbl}
\usepackage{multirow}
\usepackage{tcolorbox}

\definecolor{darkblue}{rgb}{0, 0, 0.5}
\hypersetup{colorlinks=true, citecolor=darkblue, linkcolor=darkblue, urlcolor=darkblue}

\newtcolorbox[list inside=prompt,auto counter]{prompt}[1][]{
    colbacktitle=black!60,
    coltitle=white,
    fontupper=\footnotesize,
    boxsep=5pt,
    left=2pt,
    right=2pt,
    top=2pt,
    bottom=2pt,
    boxrule=1pt,
    #1,
}

\title{Do LLM Evaluators Prefer Themselves \\ for a Reason?}

\author{Wei-Lin Chen\textsuperscript{1}
\quad
Zhepei Wei\textsuperscript{1}
\quad
Xinyu Zhu\textsuperscript{1}
\quad
Shi Feng\textsuperscript{2}
\quad
Yu Meng\textsuperscript{1}\\\\
\textsuperscript{1}University of Virginia
\quad
\textsuperscript{2}George Washington University\\\\
\small{\texttt{\{wlchen,zhepei.wei,xinyuzhu,yumeng5\}@virginia.edu}}
\quad
\small{\texttt{shi.feng@gwu.edu}}
}

\newcommand{\ie}{{\sl i.e.}}
\newcommand{\eg}{{\sl e.g.}}

\iclrfinalcopy 
\begin{document}

\maketitle

\begin{abstract}
Large language models (LLMs) are increasingly used as automatic evaluators in applications such as benchmarking, reward modeling, and self-refinement.
Prior work highlights a potential \emph{self-preference} bias where LLMs favor their own generated responses, a tendency often intensifying with model size and capability.
This raises a critical question: 
Is self-preference harmful, or does it simply reflect the genuinely higher-quality outputs of stronger models?
Answering this has been difficult because previous studies relied primarily on subjective tasks. 
These tasks lack an objective ground truth, meaning that either preference can be reasonably justified.
To address this ambiguity, we investigate self-preference using verifiable benchmarks (mathematical reasoning, factual knowledge, code generation) that allow objective ground-truth assessment. 
This enables us to distinguish \emph{harmful} self-preference (favoring objectively worse responses) from \emph{legitimate} self-preference (favoring genuinely superior ones). 
We conduct large-scale experiments under controlled evaluation conditions across diverse model families (\eg, Llama, Qwen, Gemma, Mistral, Phi, GPT, DeepSeek).
Our findings reveal three key insights:
(1)
While stronger models exhibit greater self-preference, much of this preference aligns with objectively superior performance, indicating stronger models prefer themselves mostly legitimately.
(2) Harmful self-preference persists when evaluator models err as generators, and stronger models display more pronounced harmful self-preference bias when they do err.
This suggests stronger models struggle more to recognize when they are wrong.
(3) Inference-time scaling strategies, such as generating a long Chain-of-Thought before evaluation, effectively reduce the harmful self-preference.
\textcolor{black}{Additionally, we experiment with LMArena and show that our findings extend beyond verifiable benchmarks to real-world, subjective domains.}
These results provide a more nuanced understanding of LLM-based evaluation and practical insights for improving its reliability.\footnote{Code and artifacts are available at \texttt{\href{https://github.com/wlchen0206/llm-sp}{https://github.com/wlchen0206/llm-sp}}}


\end{abstract}
\section{Introduction}
Large language models (LLMs) are increasingly adopted as automatic evaluators in various applications such as model-based benchmarking~\citep{zheng2023judging,dubois2024lengthcontrolled,fu2023gptscore, yuan2023evaluating, zengevaluating, shashidhar2023democratizing}, reward modeling~\citep{leike2018scalable,stiennon2020learning,wu2021recursively,lee2023rlaif}, self-refinement~\citep{madaan2023self,saunders2022self,shridhar2023art}, and AI oversight~\citep{bai2022constitutional,on-scalable-oversight}.
Beyond their strong alignment with human judgments~\citep{zheng2023judging}, LLM evaluators offer scalability~\citep{judgelm}, consistency~\citep{foundation-autoraters}, and cost-effectiveness~\citep{lee2023rlaif}, making them attractive for evaluating model outputs at scale.

\begin{figure}[!t]
    \centering
    \includegraphics[width=\textwidth]{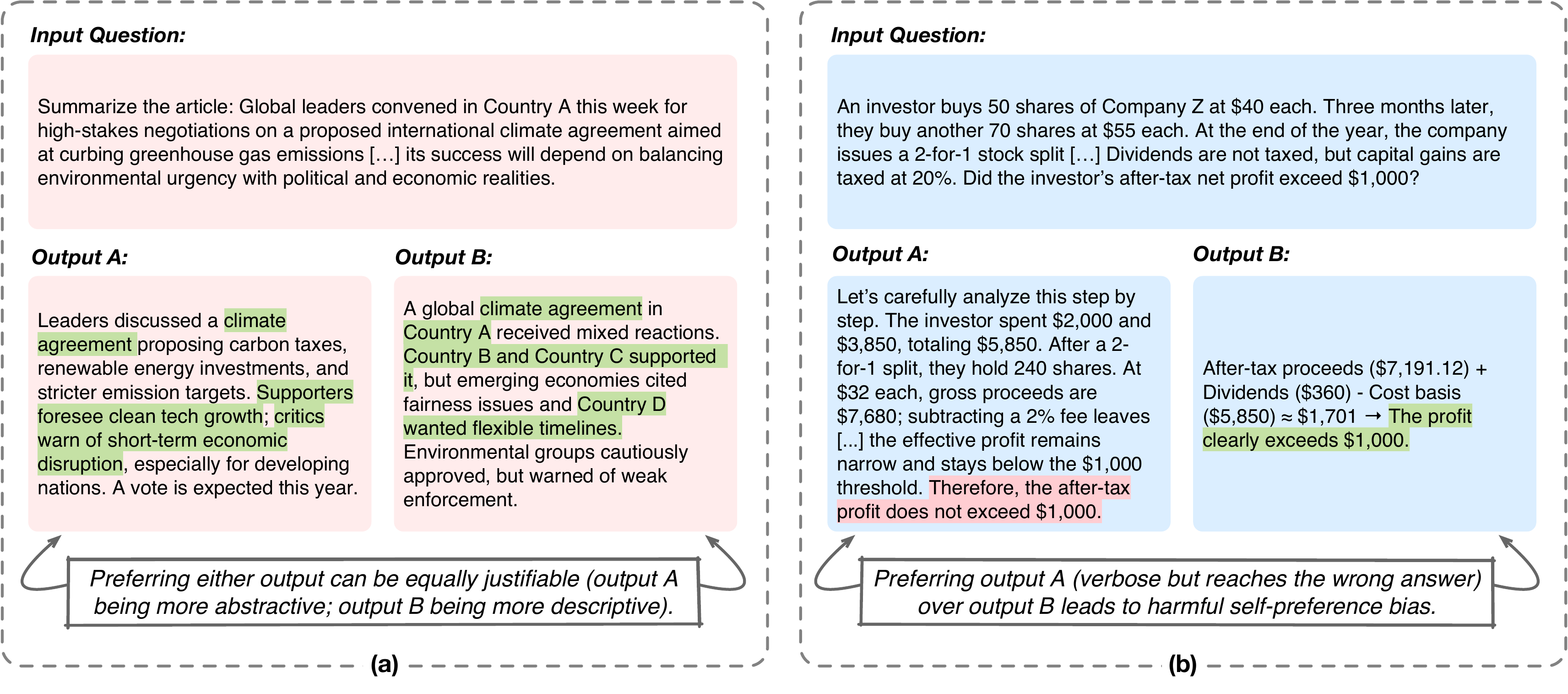} 
    \vspace{-1.5em}
    \caption{
    \textit{Left (a)}: In subjective tasks such as summarization, both outputs can be defended---one being more abstractive (Output A), the other more descriptive (Output B). Because no single ground truth exists, an evaluator’s preference remains subjective and ambiguous.
    \textit{Right (b)}: In objective tasks such as math reasoning, a unique correct answer enables clear distinction between legitimate preference and harmful bias. Here, preferring Output A (which is verbose yet incorrect) over Output B (which is concise and correct) exemplifies harmful self-preference. Such cases pose real risks for applications like LLM-as-a-Judge benchmarks, reward modeling, or self-improvement cycles, where bias can mislead evaluation and reinforce errors.
    }
    \label{fig:sub_vs_obj}
    \vspace{-1.em}
\end{figure}

However, this growing reliance on LLMs as evaluators also introduces new concerns.
One prominent issue is self-preference bias~\citep{panickssery2024llm,wataoka2024self,ye2024justice,li2024crowdsourced,liu2024llms,xu2024pride}---where LLMs exhibit a tendency to favor their own generated responses over those produced by other models.
Previous studies demonstrate this bias is typically more pronounced in larger, more capable models~\citep{zheng2023judging,li2024crowdsourced,panickssery2024llm,wataoka2024self}, raising questions about the reliability of LLM-based evaluations.
A critical question emerges: Is self-preference really a harmful bias, leading to inflated evaluations of their own outputs, or could it reflect genuine quality differences, where stronger models produce objectively superior outputs?
This question remains largely unanswered, as prior studies have typically focused on subjective and open-ended tasks such as conversational dialogue or text summarization~\citep{zheng2023judging,dubois2024lengthcontrolled,panickssery2024llm}.
In these scenarios, the lack of objective criteria for assessing the output makes it challenging to disentangle actual quality from bias (\Cref{fig:sub_vs_obj}).

In this work, we investigate self-preference using verifiable benchmarks with ground-truth references.
This setting enables us to objectively assess both LLMs’ task performance as generators and their accuracy as evaluators, thereby allowing for a clearer distinction between legitimate preference and harmful bias.
Our study spans three representative domains, including
(1) math reasoning, evaluating math word problems through numerical answer matching~\citep{hendrycks2021measuring}; 
(2) factual knowledge, assessing fact-based questions with multiple-choice answers~\citep{hendrycks2020measuring-268}; 
and (3) code generation, validating correctness through executable results~\citep{liu2023your}.

We conduct a large-scale, systematic analysis to overcome the limitations of prior work, which often
examined only a few evaluator-evaluatee pairings~\citep{zheng2023judging,li2024crowdsourced,xu2024pride} or lacked consistent cross-evaluator comparisons due to varying evaluatee sets~\citep{panickssery2024llm,wataoka2024self}.
We utilize diverse model families and sizes, including  Llama~\citep{grattafiori2024llama}, Qwen~\citep{yang2024qwen2}, Gemma~\citep{team2024gemma}, Mistral~\citep{Jiang2023Mistral7}, Phi~\citep{abdin2024phi}, GPT~\citep{hurst2024gpt}, and reasoning-enhanced long Chain-of-Thought (CoT) DeepSeek-R1 distilled models~\citep{guo2025deepseek}, as evaluators and evaluatees.
By ensuring all judge models evaluate the same set of evaluatees spanning a wide range of capabilities, our approach facilitates robust and systematic investigations of the self-preference behavior.

Our empirical findings yield three major insights:
\begin{itemize}[leftmargin=15pt]
    \item 
    While stronger models exhibit more pronounced self-preference, much of this preference aligns with objectively superior performance (\Cref{fig:corre_task_acc_sp_ratio,fig:corr_task_acc_legit_sp_ratio}).
    This suggests stronger models prefer themselves mostly legitimately.
    \item Harmful self-preference bias (\ie, favoring an objectively incorrect self-generated response) persists on instances where models perform poorly as generator.
    Moreover, while stronger models generate fewer incorrect responses overall, they show a greater tendency towards this harmful bias when they are incorrect (\Cref{fig:corre_task_acc_harmful_sp_ratio}).
    \item Employing inference-time scaling techniques for LLM evaluators (\eg, generating long CoT traces before the verdict) effectively mitigates harmful self-preference bias (\Cref{fig:reasoning_harmful_self-preference}).
\end{itemize}

\textcolor{black}{
Additionally, we also experiment on a large-scale, real-world subjective dataset sourced from LMArena~\citep{chiang2024chatbot,tang2025explorer}.
The results show that our core findings persist in subjective tasks as well, indicating that our conclusions extend beyond strictly verifiable domains.}
Lastly, we discuss how our findings provide practical recommendations for LLM-based evaluation, and explore implications for scalable oversight to address broader research interests.
\section{Experimental Setup}\label{sec:epr}
\subsection{Measuring Self-Preference in LLM-Based Evaluations}\label{sec-exp-measure}
\paragraph{LLM-based evaluation setup.}
In our experiments, we follow the pairwise evaluation format commonly used in the LLM-as-a-Judge pipeline~\citep{zheng2023judging,li2024crowdsourced,dubois2024lengthcontrolled}.
Specifically, an LLM evaluator is presented with a user query $x$ and two responses, $y_\mathcal{A}$ and $\ y_\mathcal{B}$, generated by models $\mathcal{A}$ and $\mathcal{B}$, respectively. 
The LLM evaluator $\mathcal{J}$ is instructed to act as an impartial judge and assess the quality of both responses anonymously, and provide a \emph{three-way} verdict to determine whether $y_\mathcal{A}$ is better, $y_\mathcal{B}$ is better, or if the two responses are of comparable quality (\textit{i.e.}, a tie):
\begin{equation*}
\mathcal{J}(x, y_{\mathcal{A}}, y_{\mathcal{B}}) =
\begin{cases}
y_{\mathcal{A}}, & \text{$y_\mathcal{A}$ is better}, \\
y_{\mathcal{B}}, & \text{$y_\mathcal{B}$ is better}, \\
\tau, & \text{tie}.
\end{cases}
\end{equation*}
However, prior research has shown that LLM-as-a-Judge verdicts can be sensitive to input order~\citep{zheng2023large,wang2023large-e6b,pezeshkpour2024large,wei2024unveiling,shi2024judging}.
To mitigate this position bias, we evaluate every prompt twice, swapping the order of the responses (\ie, presenting $y_\mathcal{A}$ first in one evaluation and $y_\mathcal{B}$ first in the other).
We denote the two evaluation results as $j_1 = \mathcal{J}(x, y_{\mathcal{A}}, y_{\mathcal{B}})$ and $j_2 = \mathcal{J}(x, y_{\mathcal{B}}, y_{\mathcal{A}})$. 
We define the final aggregated verdict $\mathcal{J}^*(x, y_{\mathcal{A}}, y_{\mathcal{B}})$ as follows:
\begin{equation}
\label{eq:agg_verdict}
\mathcal{J}^*(x, y_{\mathcal{A}}, y_{\mathcal{B}}) =
\begin{cases}
j_1, & \text{if } j_2 = \tau \text{ and } j_1 \neq \tau,\\
j_2, & \text{if } j_1 = \tau \text{ and } j_2 \neq \tau,\\
j_1, & \text{if } j_1 = j_2,\\
\tau, & \text{if } j_1 \neq \tau, j_2 \neq \tau, \text{and } j_1 \neq j_2.
\end{cases}
\end{equation}
Intuitively, if one evaluation yields a decisive verdict (\ie, not a tie) while the other results in a tie, we adopt the decisive outcome. 
If both evaluations agree, we return their shared verdict. 
If both are decisive but disagree, the result is a tie.
Prompts for evaluators are provided in Appendix~\ref{app:prompt}.

\paragraph{Quantifying self-preference.}
By definition, self-preference occurs when LLM evaluators favor their own generations:
Given a user prompt $x$, a judge model $\mathcal{J}$'s response $y_{\mathcal{J}}$ and another model $\mathcal{G}$'s response $y_{\mathcal{G}}$, the judge model prefers its own response $y_{\mathcal{J}}$ (i.e., $\mathcal{J}^*(x, y_{\mathcal{J}}, y_{\mathcal{G}}) = y_{\mathcal{J}}$).
Prior studies of self-preference typically examined only a small number of specific evaluator-evaluatee pairings~\citep{zheng2023judging,li2024crowdsourced,xu2024pride}, or 
used varying sets of evaluatees for different judges~\citep{panickssery2024llm,wataoka2024self}, hindering systematic cross-judge comparisons.

To establish a consistent and systematic evaluation and comparison of self-preference across diverse model families and sizes, we define two sets: 
a set of judge models $\mathcal{S}_{\mathcal{J}} = \{\mathcal{J}_1, \mathcal{J}_2, \dots, \mathcal{J}_N\}$ and a set of evaluatee models $\mathcal{S}_{\mathcal{G}} = \{\mathcal{G}_1, \mathcal{G}_2, \dots, \mathcal{G}_M\}$.
Each judge model $\mathcal{J}_i \in \mathcal{S}_{\mathcal{J}}$ evaluates its own response against those of all the evaluatee models $\mathcal{G}_k \in \mathcal{S}_{\mathcal{G}}$ in pairwise comparisons on a dataset $\mathcal{D}$. 
By keeping $\mathcal{S}_{\mathcal{G}}$ and $\mathcal{D}$ consistent across all judge models, we can systematically compare self-preference behavior across different judge model configurations, including model size, family, and capability.

To quantify self-preference, we define the \emph{self-preference ratio} (SPR) of a judge model $\mathcal{J}$ as
\begin{equation}
\label{eq:spr}
\text{SPR}_\mathcal{J} = \frac{1}{|\mathcal{S}_{\mathcal{G}}| |\mathcal{D}|} \sum_{\mathcal{G} \in \mathcal{S}_{\mathcal{G}}} \sum_{x\in \mathcal{D}} \mathbbm{1} \left\{\mathcal{J}^*(x, y_{\mathcal{J}}, y_{\mathcal{G}}) = y_{\mathcal{J}} \right\},
\end{equation}
where $\mathbbm{1}\{\cdot\}$ denotes the indicator function that returns $1$ if the condition is true.
$\text{SPR}$ represents the proportion of cases where the judge model favors its own response over those from other models using the aggregated verdict in \Cref{eq:agg_verdict}.
\Cref{eq:spr} does not distinguish between harmful and legitimate self-preference, and we will further introduce metrics for legitimate and harmful self-preference in \Cref{sec:legitimate,sec:harmful}.

\subsection{Models}\label{sec:models}
\paragraph{Evaluators.}
To evaluate how self-preference in LLMs changes across model capabilities, we construct $\mathcal{S}_{\mathcal{J}}$ by including 11 models from three representative LLM families with varying parametric scales, including:
(1) Qwen2.5 at 3B, 7B, 14B, 32B, and 72B~\citep{yang2024qwen2};
(2) Llama-3.2 at 3B, Llama-3.1 at 8B and 70B, and Llama-3.3 at 70B~\citep{grattafiori2024llama};
(3) Gemma-2 at 9B and 27B~\citep{team2024gemma}.
All models are instruction-tuned versions by default.
We also adopt long CoT reasoning models which will be discussed in Section~\ref{sec-cot-harmful}.
\paragraph{Evaluatees.}
We use a fixed set of seven evaluatee models, encompassing both weaker models and strong proprietary ones that represent a broad range of capabilities.
This set includes Llama-3.2-1B, Gemma-2-2B, Mistral-7B~\citep{Jiang2023Mistral7}, Mistral-Small, Phi-3.5~\citep{abdin2024phi}, and two proprietary models GPT-3.5-Turbo and GPT-4o.
All models are also instruction-tuned versions.
For all evaluators and evaluatees, we generate verdicts and responses in zero-shot, using greedy decoding by default.

\subsection{Tasks}
\label{sec:tasks}
\paragraph{Mathematical reasoning.}
To assess whether LLM evaluators can identify correct step-by-step solutions for math word problems, we use the MATH500 datasets~\citep{lightman2023let}, a curated subset of 500 problems selected from the full MATH dataset introduced by~\citet{hendrycks2021measuring}, with accuracy as the evaluation metric.
\paragraph{Factual knowledge.}
In addition, we employ the popular MMLU~\citep{hendrycks2020measuring-268} benchmark, evaluated by accuracy, to test whether LLM evaluators can identify accurate answers for factoid questions---a common and foundational capability in more complex queries involving world knowledge.
\paragraph{Code generation.}
We adopt the popular code generation benchmark, MBPP+~\citep{liu2023your}, which is an enhanced version of the original MBPP~\citep{austin2021program} with more robust test cases.
The results are evaluated using Pass@1.

More implementation details and prompts are provided in \Cref{app:imple,app:prompt}.
\begin{figure}[!t]
    \centering
    \includegraphics[width=\textwidth]{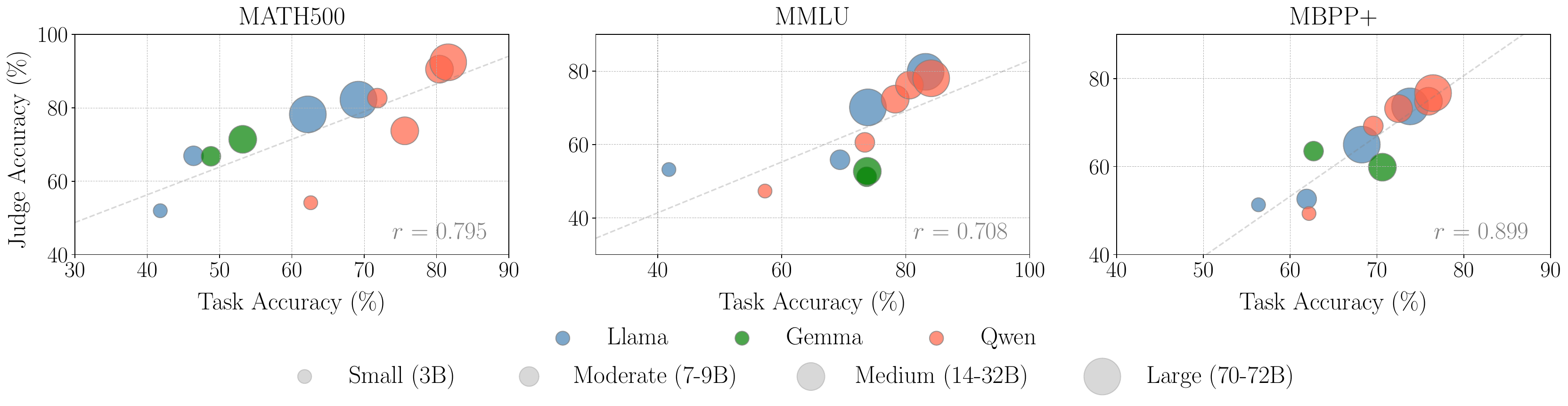} 
    \caption{
    Correlation between judge accuracy (LLMs as evaluators) and task accuracy (LLMs as generators) measured by Pearson correlation coefficient $r$~\citep{cohen2009pearson}.
    Each circle represents one LLM, with the size and color denoting model scale and family, respectively.
    The strong positive correlation between judge accuracy and task accuracy indicates that strong generators are generally accurate evaluators.
    We provide full results in \Cref{tab:full-res-jud_acc-math,tab:full-res-jud_acc-mmlu,tab:full-res-jud_acc-mbpp}.
    }
    \label{fig:corre_task_acc_judge_acc}
    \vspace{-1em}
\end{figure}

\section{Strong Models Prefer Themselves Mostly Legitimately}
\label{sec:corr}

To understand whether LLM evaluators prefer their own generations due to objective quality, we need to quantify the judge model's performance both as a generator and an evaluator.
We use the task metrics in \Cref{sec:tasks} to measure the generator performance, and define the following judge accuracy metric to evaluate the judge performance:
\begin{equation}
    \text{Judge\_Acc}_{\mathcal{J}} = \frac{1}{|\mathcal{S}_{\mathcal{G}}|} 
    \sum_{\mathcal{G} \in \mathcal{S}_{\mathcal{G}}} 
    \frac{1}{|\mathcal{D}_{\text{diff}}|}
    \sum_{x\in \mathcal{D}_{\text{diff}}} \mathbbm{1} \left\{\mathcal{J}^*(x, y_{\mathcal{J}}, y_{\mathcal{G}}) = y^* \right\},
\end{equation}
where $\mathcal{D}_{\text{diff}}$ refers to a differential  subset of $\mathcal{D}$ that includes only instances where either $y_{\mathcal{J}}$ or $y_{\mathcal{G}}$ is correct, but not both, with $y^*$ denoting the correct one.
By focusing on $\mathcal{D}_{\text{diff}}$, we ensure the judge's task is unambiguously identifying the correct response, rather than choosing based on stylistic preferences (when both are correct) or facing an ill-defined accuracy task (when both are incorrect).

\subsection{Better Generators Are Generally Better Evaluators}\label{sec-subsec-task_acc-judge_acc}


Figure~\ref{fig:corre_task_acc_judge_acc} demonstrates the correlation between task accuracy and evaluation (judge) accuracy measured with the Pearson correlation coefficient across three benchmark tasks: MATH500, MMLU, and MBPP+.
Each point represents a model, with colors distinguishing different model families (Llama, Gemma, and Qwen) and marker sizes indicating model scale.
The results reveal a clear positive correlation between task accuracy and judge accuracy, with an $r$ value of 0.795, 0.708, and 0.899 for MATH500, MMLU, and MBPP+, respectively.
Our results suggest that models capable of generating more accurate responses are also typically better evaluators.
\paragraph{Scaling and performance.}
Larger models demonstrate greater reliability as evaluators.
For example, Qwen2.5-72B and Llama-3.3-70B exhibit significantly higher evaluation accuracy compared to their smaller counterparts.
While smaller models also exhibit a positive correlation, their evaluation accuracy often plateaus at lower levels, implying that scaling improves evaluation capabilities alongside generation abilities.
\paragraph{Implications for tasks.}In mathematical reasoning, models that solve problems accurately are more adept at recognizing correct solutions from others.
Similarly, in code generation, models that produce correct code are better at identifying bugs and errors in peer code.
In factual knowledge tasks, models with higher factual accuracy are more reliable at distinguishing their correct answers from incorrect ones.
The consistent strong correlation across all three benchmarks reinforces our findings.

Overall, this relationship partly provides  justification for the reliability of LLM evaluators, especially when using the most performant, state-of-the-art models for evaluation.

\subsection{Strong Evaluators Favor Themselves, and They Are Mostly Indeed Better}
\label{sec:legitimate}

\begin{figure}[!t]
    \centering
    \includegraphics[width=\textwidth]{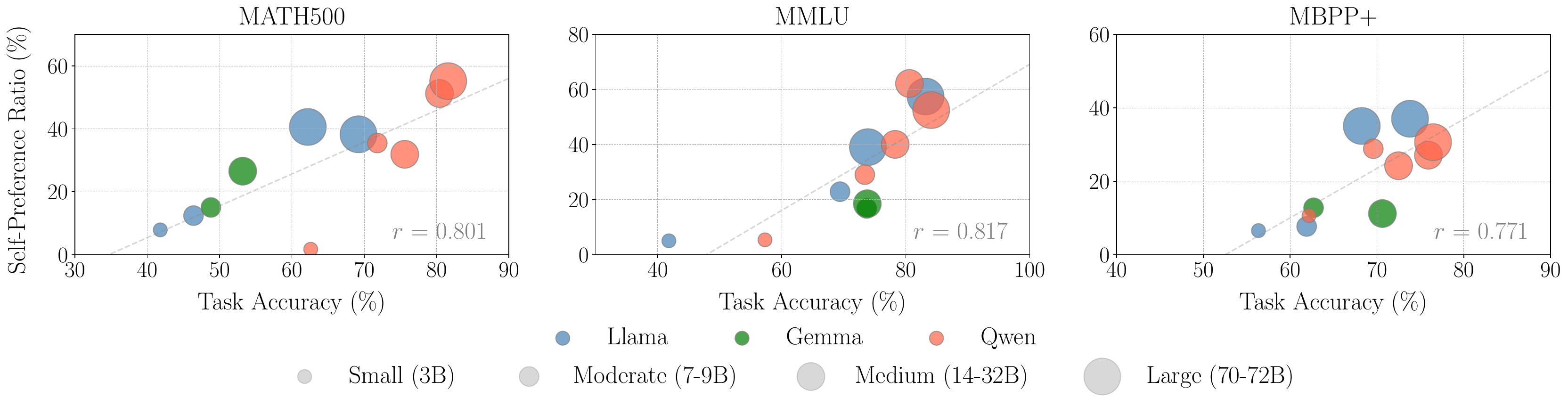} 
    \caption{Correlation between self-preference ratio and task accuracy. The clear positive correlation suggests that better generators typically prefer themselves more as evaluators. We provide full results in \Cref{tab:full-res-spr-math,tab:full-res-spr-mmlu,tab:full-res-spr-mbpp}.}
    \label{fig:corre_task_acc_sp_ratio}
  \vspace{-1em}
\end{figure}

\begin{figure}[!t]
    \centering
    \includegraphics[width=\textwidth]{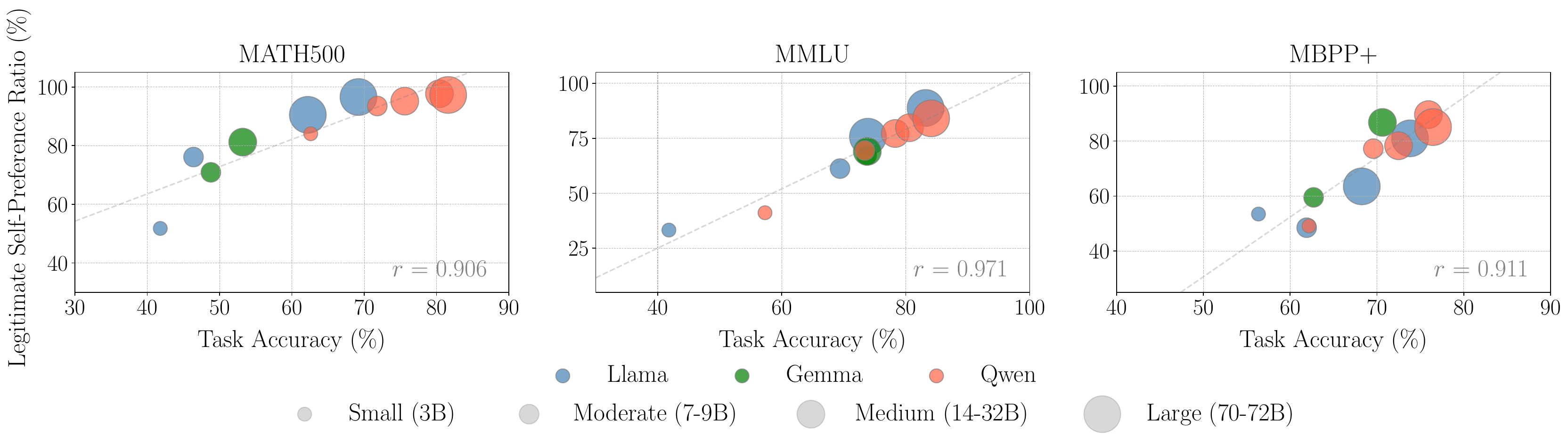} 
    \caption{Correlation between legitimate self-preference ratio and task accuracy. The consistent positive correlation indicates that when strong models favor themselves, they are mostly objectively correct. We provide full results in \Cref{tab:full-res-lspr-math,tab:full-res-lspr-mmlu,tab:full-res-lspr-mbpp}.}
  \label{fig:corr_task_acc_legit_sp_ratio}
  \vspace{-0.75em}
\end{figure}

Figure~\ref{fig:corre_task_acc_sp_ratio} illustrates the correlation between model's task accuracy as generators and self-preference ratio (\Cref{eq:spr}) as evaluators.
Similar to Section~\ref{sec-subsec-task_acc-judge_acc}, the results also reveal a clear positive correlation between task accuracy and self-preference ratio, with an $r$ value of 0.801, 0.817, and 0.771 for MATH500, MMLU, and MBPP+, respectively.
The results indicate that models with superior task performance tend to exhibit stronger self-preference.
Similarly, scaling up model size amplifies both task performance and self-preference, and this effect aligns with their superior judge accuracy.
\textcolor{black}{
To further disentangle the model's stylistic bias from its identification of objective correctness, we report SPR under $\mathcal{D}_{\text{diff}}$ (instances where either $y_{\mathcal{J}}$ or $y_{\mathcal{G}}$ is correct, but not both) and $\mathcal{D}_{\text{agree}}$ (instances where both $y_{\mathcal{J}}$ and $y_{\mathcal{G}}$ are correct or incorrect) in \Cref{fig:d_diff_vs_d_agree}.
The results demonstrate that as models become more capable, their self-preference is increasingly driven by the objective quality of the response rather than stylistic bias.
}

\paragraph{Legitimate self-preference.}
Collectively, the findings of Figure~\ref{fig:corre_task_acc_judge_acc} and~\ref{fig:corre_task_acc_sp_ratio} suggest that the higher degree of self-preference in stronger models is primarily driven by the objective quality of their outputs (evident by their higher judge accuracy), and less of the potential artifact of bias.
Since stronger models generate more accurate responses, their preference for their own outputs could be largely justified, that is, the behavior of legitimate self-preference.
We further quantified such behavior by defining \textit{legitimate self-preference ratio} (LSPR) of a judge model $\mathcal{J}$ as follows:

\begin{equation}
\text{LSPR}_{\mathcal{J}} =  \frac{1}{|\mathcal{S}_{\mathcal{G}}|} \sum_{\mathcal{G} \in \mathcal{S}_{\mathcal{G}}} 
\frac{
\sum_{x\in \mathcal{D}_{\text{diff}}}
\mathbbm{1} \left\{\mathcal{J}^*(x, y_{\mathcal{J}}, y_{\mathcal{G}}) = y_{\mathcal{J}} \text{\,  and  \,} y^* = y_{\mathcal{J}} \right\}}
{\sum_{x\in \mathcal{D}_{\text{diff}}}
\mathbbm{1} \left\{\mathcal{J}^*(x, y_{\mathcal{J}}, y_{\mathcal{G}}) = y_{\mathcal{J}} \right\}},
\end{equation}
which quantifies the degree of legitimate self-preference (\ie, $y_{\mathcal{J}}$ is preferred and correct).


As shown in Figure~\ref{fig:corr_task_acc_legit_sp_ratio}, we observe a clear trend demonstrating that an LLM’s capability as a task performer positively correlates with its legitimate self-preference ratio.
This suggests that as models become more powerful, the proportion of self-preference that is legitimate---where the model favors its own outputs when they are objectively better---also increases.
Across different model families (Llama, Gemma, and Qwen) and parameter scales, larger models consistently exhibit higher LSPR values.
In particular, Qwen-2.5-70B and Llama-3-70B achieve LSPR of 96.57$\%$ and 95.16$\%$ on MATH500, respectively.
Similarly on MBPP+ and MMLU, Qwen-2.5-70B and Llama-3-70B record an LSPR of 80.98$\% \sim$ 88.78$\%$.
Overall, the results suggest that strong models favor themselves mostly legitimately.
\textcolor{black}{We additionally experiment with a capability-matched setting using evaluator–evaluatee pairs with similar parameter scales and task accuracies (see \Cref{tab:same-scale-math-lspr,tab:same-scale-mmlu-lspr,tab:same-scale-mbpp-lspr}). Across all three datasets, we observe the same findings as in the fixed-evaluatee setting.}  
\vspace{-1.5em}




\textcolor{black}{
\section{Results on Subjective Benchmarks}\label{sec:sub}
\begin{figure}[!t]
    \centering
    \includegraphics[width=\textwidth]{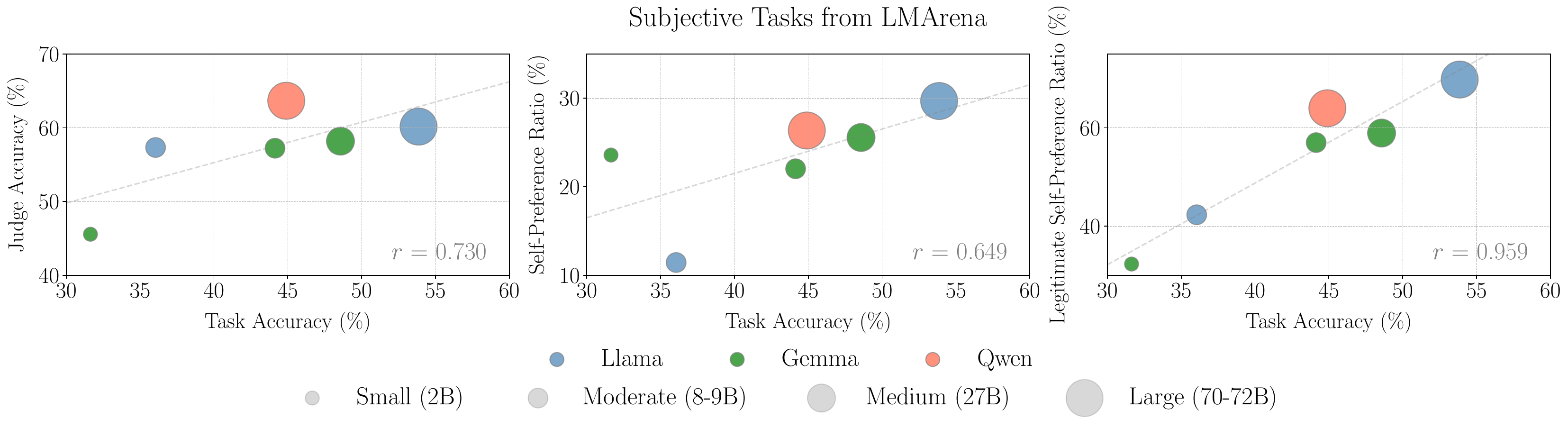} 
    \caption{\textcolor{black}{Correlation between judge accuracy/self-preference ratio/legitimate self-preference ratio and task accuracy. The consistent positive correlations suggest that our findings extend beyond verifiable tasks to real-world, subjective user queries.}}
  \label{fig:subjective-task-corre}
  \vspace{-1em}
\end{figure}
To further explore whether our findings in \Cref{sec:corr} extend beyond verifiable benchmarks to more subjective domains, we adopt a large-scale real-world dataset\footnote{\url{https://huggingface.co/datasets/lmarena-ai/arena-human-preference-100k}} collected from LMArena~\citep{chiang2024chatbot,tang2025explorer}.
Each sample of the dataset consists of a user-written query, a pair of responses produced by two different models (model $\mathcal{A}$ and model $\mathcal{B}$), and a human preference label indicating ``prefer model $\mathcal{A}$," ``prefer model $\mathcal{B}$," ``tie," or ``tie (both bad)."
This structure makes the dataset well suited for our setting, as it enables us to apply the same pairwise LLM-as-a-Judge evaluation described in \Cref{sec-exp-measure}.
The correctness of the model response is determined by the human preference label (no ground-truth or reference answers exist for these queries to verify against): a response is considered correct if the user prefers it or marks the pair as a ``tie," and incorrect if the user prefers the other response or labels the pair as ``tie (both bad)".}

\textcolor{black}{
The resulting dataset contains 12,848 samples across five evaluators:
Llama-3.1 at 8B and 70B; Gemma-2 at 2B, 9B, and 27B; and Qwen2 at 72B.
Note that the dataset is sourced from real-world user interactions, so it does not support a controlled setup with fixed queries and a fixed evaluatee pool as in \Cref{sec:corr}.
We provide the distribution of evaluatee models in \Cref{app:sub-evaluatee-distri}.
All experimental setups remain identical to those in \Cref{sec:epr} unless otherwise specified.
Additional details on constructing our subjective-task dataset from LMArena are provided in \Cref{app:sub-data}.
\paragraph{Results.}
We present results in \Cref{fig:subjective-task-corre}.
Using the correctness defined by human preference labels, we can compute task accuracies for the evaluators.
The findings and results on all metrics discussed in verifiable benchmarks remain consistent in subjective, real-world LMArean tasks.
The positive correlation between judge accuracy and task accuracy indicates that stronger generators tend to be more reliable evaluators, and similar trends on the correlation between self-preference ratio and task accuracy suggest better-performing models display stronger self-preference, an effect that increases with model scale and aligns with their improved judge accuracy.
Lastly, stronger models exhibit higher legitimate self-preference, collectively suggesting that their self-preference is typically justified by objectively superior outputs as model capability increases.
Overall, these results demonstrate that our findings would potentially generalize beyond verifiable benchmarks to subjective tasks.}

\section{Generating (Long) CoT Reduces Harmful Self-Preference}\label{sec-cot-harmful}
\subsection{Harmful Self-Preference in LLM Evaluators}
\label{sec:harmful}
While our earlier findings suggest self-preference in capable models is often benign, reflecting their genuinely higher output quality, a critical question remains regarding evaluation reliability: 
What occurs when the evaluator model $\mathcal{J}$ itself generates an objectively incorrect response $y_{\mathcal{J}}$, while an alternative response $y_{\mathcal{G}}$ is correct? 
To quantify the tendency for bias in these potentially harmful situations, we introduce the \emph{harmful self-preference propensity} (HSPP) for a judge model $\mathcal{J}$ as
\begin{equation}
\text{HSPP}_{\mathcal{J}} =  \frac{1}{|\mathcal{S}_{\mathcal{G}}|} \sum_{\mathcal{G} \in \mathcal{S}_{\mathcal{G}}} 
\frac{
\sum_{x\in \mathcal{D}_{\text{diff}}}
\mathbbm{1} \left\{\mathcal{J}^*(x, y_{\mathcal{J}}, y_{\mathcal{G}}) = y_{\mathcal{J}} \text{\,  and  \,} y^* = y_{\mathcal{G}} \right\}}
{\sum_{x\in \mathcal{D}_{\text{diff}}}
\mathbbm{1} \left\{y^* = y_{\mathcal{G}} \right\}},
\end{equation}
which characterizes the tendency of an LLM evaluator to prefer its own incorrect generation over objectively better ones.
A higher HSPP indicates a greater tendency towards harmful bias when the evaluator itself has erred.

\begin{figure}[tbp]
    \centering
    \includegraphics[width=\textwidth]{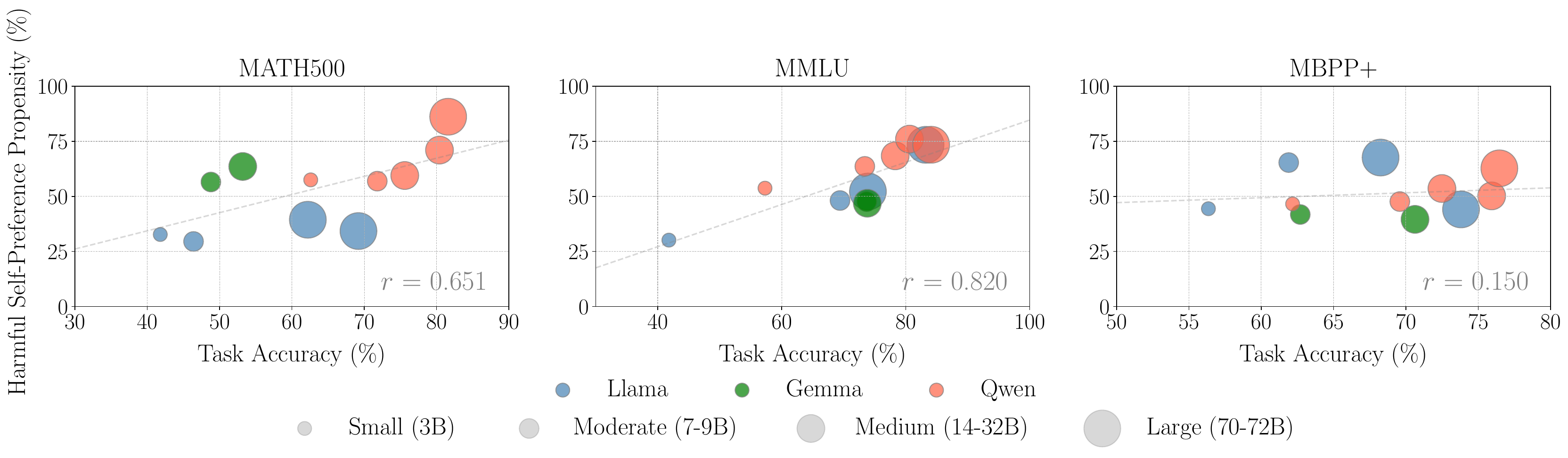} 
    \caption{Correlation between harmful self-preference propensity and task accuracy. The positive correlation, in particular on MATH500 and MMLU, implies that when strong models are objectively incorrect, they prefer themselves more often. We provide full results in \Cref{tab:full-res-hspr-math,tab:full-res-hspr-mmlu,tab:full-res-hspr-mbpp}.}
    \label{fig:corre_task_acc_harmful_sp_ratio}
    \vspace{-1em}
\end{figure}

\paragraph{Results.}
As observed in Figure~\ref{fig:corre_task_acc_harmful_sp_ratio}, when evaluators' responses are objectively worse, there exists a clear positive correlation between task performance and the harmful self-preference propensity.
In other words, stronger models---those with a higher task accuracy---tend to exhibit greater harmful self-preference when evaluating cases where their own outputs are incorrect but the alternative response is correct.
This trend is particularly pronounced in larger models.
Notably, the most performant model, Qwen2.5-72B, exhibits an HSPP of 86$\%$ on MATH500 and 73$\%$ on MMLU, significantly higher than its overall SPR (\Cref{fig:corre_task_acc_sp_ratio}) of 55$\%$ and 52$\%$, respectively.
Although MBPP+ shows a weaker positive correlation,
large models like Llama-3.1-70B and Qwen2.5-72B still favor themselves more than smaller ones, with HSPP values ranging from approximately 50$\%$ to 75$\%$, significantly higher than their overall SPR (\Cref{fig:corre_task_acc_sp_ratio}), which remains below 40$\%$.

In sum, such model behavior presents a potential safety challenge: as models become more capable, they are also more confident in their own responses, even when they are wrong.
This overconfidence could lead to biased evaluation frameworks where stronger models dismiss superior responses from other models, potentially reinforcing flawed or suboptimal outputs for AI oversight.

\begin{figure}[!t]
    \centering
    \includegraphics[width=0.8\textwidth]{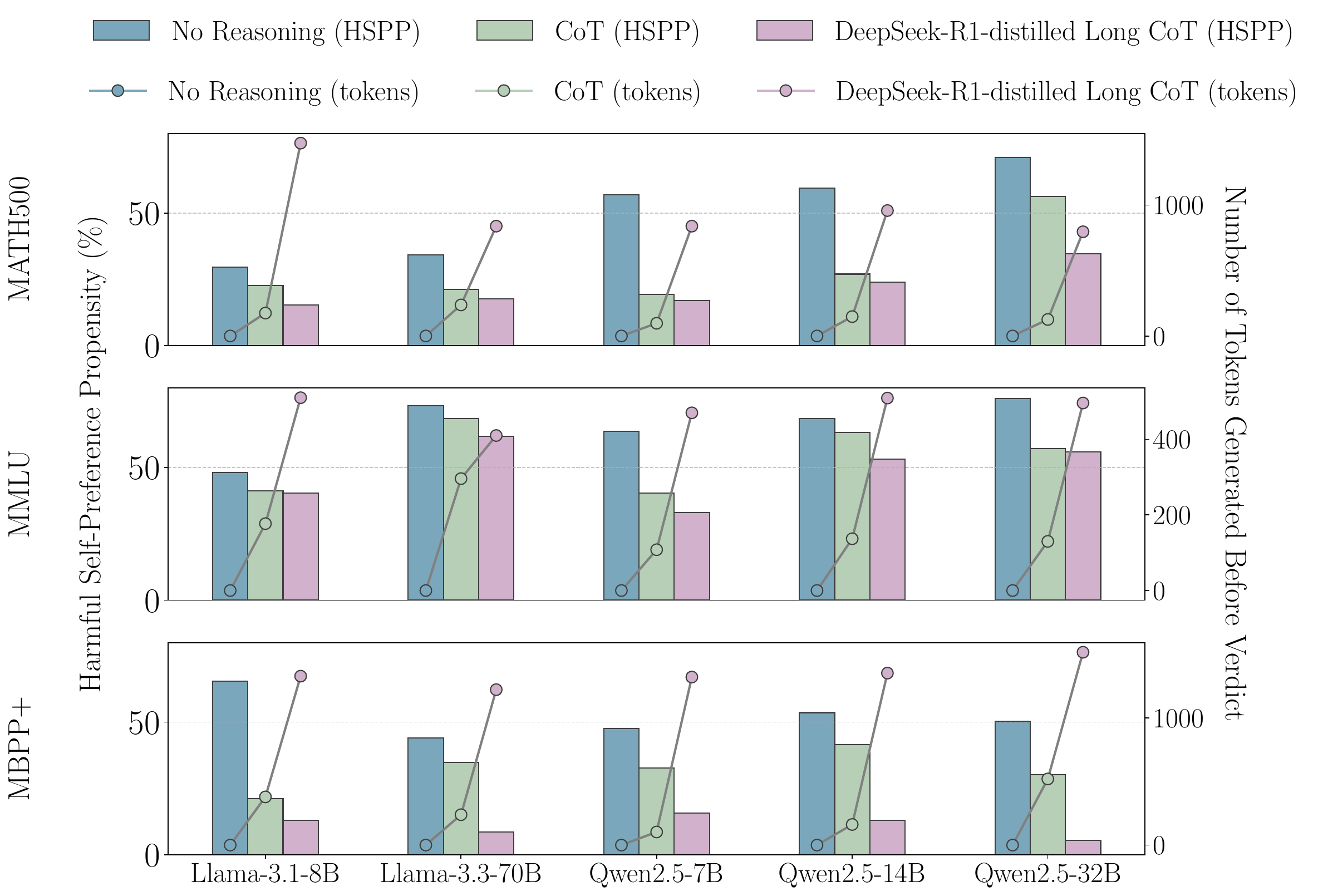} 
    \caption{The harmful self-preference propensity at varying levels of evaluator reasoning.
    When being objectively incorrect, evaluators can achieve more accurate verdict by generating more reasoning tokens prior to their judgments. We provide full results in \Cref{tab:full-res-hspp-math-reasoning,tab:full-res-hspp-mmlu-reasoning,tab:full-res-hspp-mbpp-reasoning,tab:full-res-hspp-math-reasoning-token,tab:full-res-hspp-mmlu-reasoning-token,tab:full-res-hspp-mbpp-reasoning-token}.}
    \label{fig:reasoning_harmful_self-preference}
    \vspace{-0.75em}
\end{figure}

\subsection{Mitigating Harmful Self-Preference Bias via CoT}
Compared to recognizing accurate responses, identifying its own mistakes and overriding its initial understanding may require a deeper level of analysis and reasoning for LLM evaluators.
With recent advances in reasoning-enhanced, long CoT models, we aim to investigate the impact of generating reasoning traces for mitigating harmful self-preference.
To this end, we experiment with three LLM evaluator settings: no reasoning tokens, standard CoT reasoning, and long CoT reasoning.
\paragraph{No reasoning.}
The same setting adopted in our previous experiments, where the evaluator is instructed to directly output the corresponding verdict label.
\paragraph{Standard CoT reasoning.}
The evaluator generates a step-by-step reasoning chain before arriving at its verdict.
Both no-reasoning-token setup and standard-CoT-reasoning settings adopt the same underlying instruction-tuned model.
\paragraph{Long CoT reasoning.}
In this setting, we employ 5 distilled reasoning models with matching backbones, including DeepSeek-R1-Distill-Llama-8B/70B and DeepSeek-R1-Distill-Qwen-7B/14B/32B, from DeepSeek-R1~\citep{guo2025deepseek}.
The reasoning model evaluator naturally generates a detailed multi-step reasoning trace before deriving at its verdict.
\paragraph{Results.}
The results are presented in Figure~\ref{fig:reasoning_harmful_self-preference}.
As shown on both MATH500 and MMLU, generating reasoning traces substantially reduces harmful self-preference across all models.
The no-reasoning-token setting exhibits the highest HSPP, and introducing CoT reasoning mitigate the issue to a noticeable degree, lowering harmful self-preference.
Reasoning-enhanced models with long CoT further amplify this mitigation, consistently achieving the lowest HSPP across all models.

The trend suggests that reasoning-enhanced evaluations encourage models to more accurately reassess their initial understanding and consider alternative responses more carefully.
Interestingly, the relative reduction in HSPP is more pronounced on MATH500 and MBPP+ compared to those on MMLU, possibly indicating that reasoning-intensive tasks benefit more from reasoning-driven evaluations.
Overall, these findings underscore the potential of reasoning-enhanced evaluation strategies to improve the reliability of LLM evaluators.
\section{Related Work}

\paragraph{LLM-based evaluations.}
Extensive works have explored leveraging LLMs not only as generators, but also as evaluators~\citep{llm-eval,judgelm,ankner2024critique,wei2025instructrag}.
Early work like LLM-Eval~\citep{llm-eval} introduces a unified approach that prompts an LLM to assess multiple quality dimensions of open-domain conversations, reducing costly human annotations.
JudgeLM~\citep{judgelm} further propose fine-tuning LLMs specifically to act as judges, and demonstrates that with appropriate training, LLMs can serve as scalable judges and achieve a high agreement rate with human judgments.
More recently,~\cite{liu2025inferencetimescalinggeneralistreward} investigate improving reward modeling for
general queries by scaling inference compute, and propose DeepSeek-GRM.
Prior works~\citep{lin2024criticbench,lan2024criticeval} have also observed a strong correlation between model capabilities as generators and as critics, aligning with our findings.
Their setup differs from ours in that they evaluate a fixed set of responses, whereas we focus on a pairwise judgment setting in which one of the evaluatees is always the same as the evaluator.
However, it is observed that LLM-based evaluators are often biased~\citep{zheng2023judging, dubois2024lengthcontrolled, foundation-autoraters, great-model-think-alike, on-scalable-oversight}.
For example,~\citet{dubois2024lengthcontrolled} employ a regression-based control mechanism to mitigate the length bias prevalent in LLM evaluators.
\citet{foundation-autoraters} reduce response ordering bias and length bias in LLM evaluators by training on a diverse set of quality assessment tasks.
In this work, we present an in-depth analysis of self-preference bias in LLMs and provide empirical insights for improving the reliability of LLM-based evaluations.



\paragraph{Self-preference bias in LLM evaluators.}

Several studies have highlighted a prevalent \emph{self-preference bias}~\citep{bai2023benchmarking, li2024crowdsourced, ye2024justice}, also known as self-enhancement bias~\citep{zheng2023judging}), where LLM evaluators favor their own generations and potentially lead to biased evaluation~\citep{bitton2023visit, liu2024llms}.
\citet{koo2024benchmarking} examine this from a cognitive aspect; other works~\citep{xu2024pride,stureborg2024large} propose quantified metrics (\eg, scalar scores) to measure the degree of self-preference bias.
Recently,~\citet{panickssery2024llm} discover a linear correlation between self-recognition capability and the strength of self-preference, and~\citet{li2025preference} explore the preference leakage problem---whether evaluators prefer outputs from student models distilled from teacher models that are similar to the evaluators.
However, many of them have either constrained their analysis to a limited number of models~\citep{zheng2023judging, li2024crowdsourced, xu2024pride} or evaluated LLMs under setups where evaluators and evaluatees are drawn from the same model pool~\citep{panickssery2024llm, wataoka2024self}, resulting in different evaluators being applied to different sets of evaluatees and obscuring the underlying causes of bias.
In contrast, we introduce a controlled setup with designated sets of evaluators and evaluatees, ensuring a consistent assessment across models.
This enables us to conduct a comprehensive, large-scale analysis across diverse model families and sizes, yielding more generalizable and systematic insights into the potential drivers and mitigations of the (harmful) self-preference.
\textcolor{black}{A few works have begun exploring potential mitigation strategies for self-preference bias.
\citet{roytburg2025breaking} introduces activation-based methods using steering vectors to suppress self-preference, though these approaches can suffer from instability.
\citet{dorner2024limits} provides a more theoretical analysis showing that when a judge is not more accurate than the model it evaluates, no debiasing method can reduce the need for ground-truth labels by more than a factor of two.
Collectively with our long-CoT mitigation results, these findings underscore both the promise and the limitations of current safeguards for LLM-as-judge and motivate the development of more robust LLM-based evaluators.}

\section{Discussions}
\label{sec:disc}
\paragraph{Practical recommendations for robust LLM-based evaluation.}

Our findings offer several practical recommendations for more reliable real-world LLM-based evaluation systems. 
First, before deploying an LLM as an evaluator for a specific task or domain, its generative performance
on the target domain should be rigorously assessed. 
This pre-assessment helps select models that are more likely to provide accurate judgments and exhibit more legitimate (rather than harmful) preference. 
Second, our results highlight two complementary pathways to mitigate harmful self-preference bias: (1) Model scaling, which employs larger, more capable evaluators. 
This primarily boosts reliability by improving generative performance, thus shrinking the set of instances where the model fails and harmful bias can occur. 
(2) Implementing inference-time scaling (\eg, CoT reasoning)
effectively reduce the propensity for harmful bias on those remaining error instances.
Finally, for large-scale systems covering diverse domains,
consider deploying a roster of specialized LLM evaluators.
Evaluation requests could then be dynamically routed to the model with the highest generative capability (and thus, likely highest evaluation accuracy) in the specific domain.
\paragraph{Implications for scalable oversight.}

A key motivation for LLM-based evaluation is achieving \emph{scalable oversight}~\citep{bai2022constitutional,bowman2022measuring,on-scalable-oversight,great-model-think-alike}---automating evaluation pipelines for complex domains like competition math or advanced coding, where human evaluation is costly or requires specialized expertise.
Before deploying LLMs in such critical roles, their reliability must be thoroughly validated.
However, the inherent complexity in these tasks that necessitates AI oversight also makes obtaining human-annotated ground truth for validation exceptionally challenging. 
Our finding that ``better generators are better evaluators'' on simpler yet still verifiable benchmarks provides a crucial bridge for establishing trust:
We can have greater confidence in evaluations from models known to excel as generators, particularly in their areas of strength, potentially reducing the need for exhaustive human review.
Complementing this, our finding that harmful self-preference bias persists, particularly when models fail on specific task instances, enables the strategic targeting of human intervention.
We may focus review efforts on evaluations from models known to struggle with certain task types, or on particularly challenging instances where self-preference bias is most likely to emerge. 
This targeted approach enables more efficient allocation of limited human review resources compared to uniform human oversight, making the entire evaluation system more scalable while maintaining reliability.
\section{Conclusion}
In this work, we investigate self-preference in LLMs using verifiable benchmarks with objective ground-truth reference.
Our experiments suggest that self-preference often reflects genuine output quality in more capable models. Stronger models tend to be more accurate judges, with their preferences frequently aligned with objective correctness.
However, harmful self-preference persists when models favor an objectively incorrect response by themselves.
To address this, we explore inference-time scaling strategies and demonstrate that, while
being objectively worse as generators, evaluators can achieve more accurate verdict by generating more reasoning tokens.
Overall, our study provides deeper insights into understanding and improving LLM-based evaluation.
We discuss the limitations of our work in \Cref{sec:limit}.

\bibliography{iclr2026_conference}
\bibliographystyle{iclr2026_conference}

\newpage
\appendix

\section{Implementation Details}
\label{app:imple}
\paragraph{Models.}
We adopt the following model checkpoints/versions from Huggingface~\citep{wolf2019huggingface} and OpenAI API\footnote{\url{https://platform.openai.com/docs/models}} for our experiments:
\begin{itemize}[leftmargin=25pt]
    \item Llama family:
    \href{https://huggingface.co/meta-llama/Llama-3.2-1B-Instruct}{meta-llama/Llama-3.2-1B-Instruct},
    \href{https://huggingface.co/meta-llama/Llama-3.2-3B-Instruct}{meta-llama/Llama-3.2-3B-Instruct},
    \href{https://huggingface.co/meta-llama/Llama-3.1-8B-Instruct}{meta-llama/Llama-3.1-8B-Instruct},
    \href{https://huggingface.co/meta-llama/Llama-3.1-70B-Instruct}{meta-llama/Llama-3.1-70B-Instruct},
    \href{https://huggingface.co/meta-llama/Llama-3.3-70B-Instruct}{meta-llama/Llama-3.3-70B-Instruct}
    \item Gemma family:
    \href{https://huggingface.co/google/gemma-2-2b-it}{google/gemma-2-2b-it},
    \href{https://huggingface.co/google/gemma-2-9b-it}{google/gemma-2-9b-it},
    \href{https://huggingface.co/google/gemma-2-27b-it}{google/gemma-2-27b-it}
    \item Qwen family:
    \href{https://huggingface.co/Qwen/Qwen2.5-3B-Instruct}{Qwen/Qwen2.5-3B-Instruct},
    \href{https://huggingface.co/Qwen/Qwen2.5-7B-Instruct}{Qwen/Qwen2.5-7B-Instruct},
    \href{https://huggingface.co/Qwen/Qwen2.5-14B-Instruct}{Qwen/Qwen2.5-14B-Instruct},
    \href{https://huggingface.co/Qwen/Qwen2.5-32B-Instruct}{Qwen/Qwen2.5-32B-Instruct},
    \href{https://huggingface.co/Qwen/Qwen2.5-72B-Instruct}{Qwen/Qwen2.5-72B-Instruct}
    \item DeepSeek-R1-Distill family:
    \href{https://huggingface.co/deepseek-ai/DeepSeek-R1-Distill-Llama-8B}{deepseek-ai/DeepSeek-R1-Distill-Llama-8B},
    \href{https://huggingface.co/deepseek-ai/DeepSeek-R1-Distill-Llama-70B}{deepseek-ai/DeepSeek-R1-Distill-Llama-70B},
    \href{https://huggingface.co/deepseek-ai/DeepSeek-R1-Distill-Qwen-7B}{deepseek-ai/DeepSeek-R1-Distill-Qwen-7B},
    \href{https://huggingface.co/deepseek-ai/DeepSeek-R1-Distill-Qwen-14B}{deepseek-ai/DeepSeek-R1-Distill-Qwen-14B},
    \href{https://huggingface.co/deepseek-ai/DeepSeek-R1-Distill-Qwen-32B}{deepseek-ai/DeepSeek-R1-Distill-Qwen-32B}
    \item Mistral family:
    \href{https://huggingface.co/mistralai/Mistral-7B-Instruct-v0.3}{mistralai/Mistral-7B-Instruct-v0.3},
    \href{https://huggingface.co/mistralai/Mistral-Small-Instruct-2409}{mistralai/Mistral-Small-Instruct-2409}
    \item Phi family:
    \href{https://huggingface.co/microsoft/Phi-3.5-mini-instruct}{microsoft/Phi-3.5-mini-instruct}
    \item GPT family:
    \href{https://platform.openai.com/docs/models/gpt-3.5-turbo}{gpt-3.5-turbo-0125},
    \href{https://platform.openai.com/docs/models/gpt-4o}{gpt-4o-2024-11-20}
\end{itemize}

\paragraph{Response generation.}
For generating responses, we adopt \texttt{temperature} = 0 (\ie, greedy decoding) for all instruction-tuned models, including OpenAI API models;
we adopt \texttt{temperature} = 0.6 for all DeepSeek-R1-distilled reasoning models as recommended in the model document to prevent endless repetitions or incoherent outputs.
Also, to prevent potential length biases in evaluator, for reasoning models we preserve only the responses after the ``$<$\textbackslash think$>$" token as the answer to be evaluated (\ie, the long, verbose reasoning traces within ``$<$think$>$" and ``$<$\textbackslash think$>$" are not presented to the evaluators).
Note that the responses after the ``$<$\textbackslash think$>$" token still contain concise explanations and rationales similar to the response from instruction-tuned, non-reasoning models.

\paragraph{Verdict generation.}
For generating verdicts, we also adopt \texttt{temperature} = 0 for all non-reasoning models and \texttt{temperature} = 0.6 for reasoning models.
By default, our experiments employ \texttt{max\_tokens} = 1, and we obtain the verdict by instructing the model to directly output a label: ``A'', ``T", or ``B", corresponding to which assistant's answer is better or they are relatively the same in quality.
Specifically, we examine the label set for their corresponding candidate tokens and select the token among the three with the highest assigned logit as the evaluator's verdict.

For experiments in Section~\ref{sec-cot-harmful}, the no reasoning token setting adopts the above-described configuration.
For the standard CoT reasoning setting, the model is allowed to freely generate its judgment in natural language with a reasoning chain, before ending its response with a verdict (\eg, ``My final verdict is $\{$verdict$\}$").
The verdict is then parsed into the corresponding label.
For long CoT reasoning, the model is instructed in the same way as the standard CoT reasoning setting, allowing it to freely generate the judgment.

\paragraph{Other details.}
We perform all model inference (except proprietary OpenAI API models) using the vLLM library~\citep{kwon2023efficient}.
The evaluation script for task accuracy is adopted from \href{https://github.com/EleutherAI/lm-evaluation-harness/blob/main/lm_eval/tasks/minerva_math/README.md}{lm-evaluation-harness} and~\citet{lewkowycz2022solving} for MATH500;
\href{https://github.com/huggingface/evaluate/tree/main/metrics/code_eval}{code\_eval} and~\citet{chen2021evaluating} for MBPP+.
We calculate the token length in \Cref{fig:reasoning_harmful_self-preference} using GPT-2 tokenizer from~\href{https://github.com/openai/tiktoken}{tiktoken}.
The majority of our experiments are conducted on cloud computing infrastructure with access to 8 NVIDIA A100 GPUs per instance.

We randomly sample 1K instances from the full MMLU test set for our MMLU experiments to ensure computational efficiency, given our large set of evaluators and evaluatees.
Our preliminary studies indicate that such a 1K subset is sufficient to produce stable results, and further expansion has negligible impact on the outcomes.

\newpage
\section{Prompts}
\label{app:prompt}

\begin{table*}[!h]
\caption{Evaluator prompt (no reasoning) for MATH500.}\label{tab:prompt-judge-no_reason-math}
\begin{prompt}[title={}, label=]
You will be given answers from two assistants (Assistant A and Assistant B) to a user prompt. Please act as an impartial judge and evaluate the quality of their answers.\\

You must output only one of the following choices:\\
  
- ``A" if Assistant A's answer is better.\\
- ``T" if both answers are relatively the same in quality.\\
- ``B" if Assistant B's answer is better.\\

For example, if both answers are relatively the same in quality, output ``T".\\

$<$User Prompt$>$\\
$\{$question 1$\}$\\

$<$The Start of Assistant A's Answer$>$\\
$\{$answer 1$\}$\\
$<$The End of Assistant A's Answer$>$\\
  
$<$The Start of Assistant B's Answer$>$\\
$\{$answer 2$\}$\\
$<$The End of Assistant B's Answer$>$
\end{prompt}
\end{table*}

\begin{table*}[!h]
\caption{Evaluator prompt (CoT) for MATH500.}\label{tab:prompt-judge-reasoning-math}
\begin{prompt}[title={}, label=]
You will be given answers from two assistants (Assistant A and Assistant B) to a user prompt. Please act as an impartial judge and evaluate the quality of their answers.\\

You must end your response with: "My final verdict is \$\$…\$\$.”, where \$\$…\$\$ must enclose one of the following:\\
  
- ``A" if Assistant A's answer is better.\\
- ``T" if both answers are relatively the same in quality.\\
- ``B" if Assistant B's answer is better.\\

For example, if your final verdict is a tie, end your response with: ``My final verdict is \$\$T\$\$."\\

$<$User Prompt$>$\\
$\{$question 1$\}$\\

$<$The Start of Assistant A's Answer$>$\\
$\{$answer 1$\}$\\
$<$The End of Assistant A's Answer$>$\\
  
$<$The Start of Assistant B's Answer$>$\\
$\{$answer 2$\}$\\
$<$The End of Assistant B's Answer$>$
\end{prompt}
\end{table*}

\begin{table*}[!h]
\caption{Evaluator prompt (long CoT) for MATH500.}\label{tab:prompt-judge-long-cot-reasoning-math}
\begin{prompt}[title={}, label=]
You will be given answers from two assistants (Assistant A and Assistant B) to a user prompt. Please act as an impartial judge and evaluate the quality of their answers.\\

You must end your response with: "My final verdict is \$\$…\$\$.”, where \$\$…\$\$ must enclose one of the following:\\
  
- ``A" if Assistant A's answer is better.\\
- ``T" if both answers are relatively the same in quality.\\
- ``B" if Assistant B's answer is better.\\

For example, if your final verdict is a tie, end your response with: ``My final verdict is \$\$T\$\$."\\

$<$User Prompt$>$\\
$\{$question 1$\}$\\

$<$The Start of Assistant A's Answer$>$\\
$\{$answer 1$\}$\\
$<$The End of Assistant A's Answer$>$\\
  
$<$The Start of Assistant B's Answer$>$\\
$\{$answer 2$\}$\\
$<$The End of Assistant B's Answer$>$\\

$<$think$>$
\end{prompt}
\end{table*}

\begin{table*}[!h]
\caption{Evaluator prompt (no reasoning) for MMLU.}\label{tab:prompt-judge-no_reason-mmlu}
\begin{prompt}[title={}, label=]
You will be given answers from two assistants (Assistant A and Assistant B) to a user prompt. Please act as an impartial judge and evaluate the quality of their answers.\\

You must output only one of the following choices:\\
  
- ``A" if Assistant A's answer is better.\\
- ``T" if both answers are relatively the same in quality.\\
- ``B" if Assistant B's answer is better.\\

For example, if both answers are relatively the same in quality, output ``T".\\

$<$User Prompt$>$\\
$\{$question 1$\}$\\

$<$The Start of Assistant A's Answer$>$\\
$\{$answer 1$\}$\\
$<$The End of Assistant A's Answer$>$\\
  
$<$The Start of Assistant B's Answer$>$\\
$\{$answer 2$\}$\\
$<$The End of Assistant B's Answer$>$
\end{prompt}
\end{table*}

\begin{table*}[!h]
\caption{Evaluator prompt (CoT) for MMLU.}\label{tab:prompt-judge-reasoning-mmlu}
\begin{prompt}[title={}, label=]
You are given answers provided by two assistants, Assistant A and Assistant B, to the following user prompt.\\

$<$User Prompt$>$\\
$\{$question 1$\}$\\

$<$The Start of Assistant A's Answer$>$\\
$\{$answer 1$\}$\\
$<$The End of Assistant A's Answer$>$\\
  
$<$The Start of Assistant B's Answer$>$\\
$\{$answer 2$\}$\\
$<$The End of Assistant B's Answer$>$\\

Please act as an impartial judge and evaluate the quality of their answers.\\

You must end your response with: "My final verdict is \$\$…\$\$.”, where \$\$…\$\$ must enclose one of the following:\\
  
- ``A" if Assistant A's answer is better.\\
- ``T" if both answers are relatively the same in quality.\\
- ``B" if Assistant B's answer is better.\\

For example, if your final verdict is a tie, end your response with: ``My final verdict is \$\$T\$\$."
\end{prompt}
\end{table*}

\begin{table*}[!h]
\caption{Evaluator prompt (long CoT) for MMLU.}\label{tab:prompt-judge-long-cot-reasoning-mmlu}
\begin{prompt}[title={}, label=]
You are given answers provided by two assistants, Assistant A and Assistant B, to the following user prompt.\\

$<$User Prompt$>$\\
$\{$question 1$\}$\\

$<$The Start of Assistant A's Answer$>$\\
$\{$answer 1$\}$\\
$<$The End of Assistant A's Answer$>$\\
  
$<$The Start of Assistant B's Answer$>$\\
$\{$answer 2$\}$\\
$<$The End of Assistant B's Answer$>$\\

Please act as an impartial judge and evaluate the quality of their answers.\\

You must end your response with: "My final verdict is \$\$…\$\$.”, where \$\$…\$\$ must enclose one of the following:\\
  
- ``A" if Assistant A's answer is better.\\
- ``T" if both answers are relatively the same in quality.\\
- ``B" if Assistant B's answer is better.\\

For example, if your final verdict is a tie, end your response with: ``My final verdict is \$\$T\$\$."\\

$<$think$>$
\end{prompt}
\end{table*}

\begin{table*}[!h]
\caption{Evaluator prompt (no reasoning) for MBPP+.}\label{tab:prompt-judge-no_reason-mbpp}
\begin{prompt}[title={}, label=]
Please act as an impartial judge and assess the accuracy of the given Python functions. Provide your evaluation by outputting only one of the following\\
  
- ``A" if Assistant A's function is more accurate.\\
- ``T" if both functions are similarly accurate.\\
- ``B" if Assistant B's function is more accurate.\\

Base your judgment on your understanding of the task described in the user prompt and whether the functions successfully pass the provided test cases.\\

$<$User Prompt$>$\\
$\{$question 1$\}$\\

$<$The Start of Assistant A's Function$>$\\
$\{$answer 1$\}$\\
$<$The End of Assistant A's Function$>$\\
  
$<$The Start of Assistant B's Function$>$\\
$\{$answer 2$\}$\\
$<$The End of Assistant B's Function$>$
\end{prompt}
\end{table*}

\begin{table*}[!h]
\caption{Evaluator prompt (CoT) for MBPP+.}\label{tab:prompt-judge-reasoning-mbpp}
\begin{prompt}[title={}, label=]
Please act as an impartial judge and assess the accuracy of the given Python functions. You must end your response with: "My final verdict is \$\$…\$\$.”, where \$\$…\$\$ must enclose one of the following:\\
  
- ``A" if Assistant A's function is more accurate.\\
- ``T" if both functions are similarly accurate.\\
- ``B" if Assistant B's function is more accurate.\\

For example, if your final verdict is a tie, end your response with: ``My final verdict is \$\$T\$\$."

Base your judgment on your understanding of the task described in the user prompt and whether the functions successfully pass the provided test cases.\\

$<$User Prompt$>$\\
$\{$question 1$\}$\\

$<$The Start of Assistant A's Function$>$\\
$\{$answer 1$\}$\\
$<$The End of Assistant A's Function$>$\\
  
$<$The Start of Assistant B's Function$>$\\
$\{$answer 2$\}$\\
$<$The End of Assistant B's Function$>$
\end{prompt}
\end{table*}

\begin{table*}[!h]
\caption{Evaluator prompt (long CoT) for MBPP+.}\label{tab:prompt-judge-long-cot-reasoning-mbpp}
\begin{prompt}[title={}, label=]
Please act as an impartial judge and assess the accuracy of the given Python functions. You must end your response with: "My final verdict is \$\$…\$\$.”, where \$\$…\$\$ must enclose one of the following:\\
  
- ``A" if Assistant A's function is more accurate.\\
- ``T" if both functions are similarly accurate.\\
- ``B" if Assistant B's function is more accurate.\\

For example, if your final verdict is a tie, end your response with: ``My final verdict is \$\$T\$\$."

Base your judgment on your understanding of the task described in the user prompt and whether the functions successfully pass the provided test cases.\\

$<$User Prompt$>$\\
$\{$question 1$\}$\\

$<$The Start of Assistant A's Function$>$\\
$\{$answer 1$\}$\\
$<$The End of Assistant A's Function$>$\\
  
$<$The Start of Assistant B's Function$>$\\
$\{$answer 2$\}$\\
$<$The End of Assistant B's Function$>$\\

$<$think$>$
\end{prompt}
\end{table*}

\begin{table*}[!h]
\caption{Generator prompt (instruction-tuned model) for MATH500.}
\begin{prompt}[title={}, label=]
Answer the given problem by providing your solution. You must use \textbackslash boxed$\{$...$\}$ to enclose the final answer.\\

Problem:\\
$\{$question$\}$\\

Solution:
\end{prompt}
\label{tab:prompt-answer-instruct-math}
\end{table*}

\begin{table*}[!h]
\caption{Generator prompt (reasoning model) for MATH500.}
\label{tab:prompt-answer-reasoning-math}
\begin{prompt}[title={}, label=]
Answer the given problem by providing your solution. You must use \textbackslash boxed$\{$...$\}$ to enclose the final answer.\\

Problem:\\
$\{$question$\}$\\

Solution:\\
$<$think$>$
\end{prompt}
\end{table*}

\begin{table*}[!h]
\caption{Generator prompt (instruction-tuned model) for MMLU.}\label{tab:prompt-answer-instruct-mmlu}
\begin{prompt}[title={}, label=]
Answer the given multiple-choice problem. If you cannot determine the correct answer, take your best guess.\\
You must end your response with "The final answer is \$\$...\$\$.", where \$\$...\$\$ must only enclose the label of your final answer. For example, if your final answer is K, then write "The final answer is \$\$K\$\$.".

Problem:\\
$\{$question$\}$\\

Solution:
\end{prompt}
\end{table*}

\begin{table*}[!h]
\caption{Generator prompt (reasoning model) for MMLU.}\label{tab:prompt-answer-reasoning-mmlu}
\begin{prompt}[title={}, label=]
Answer the given multiple-choice problem. If you cannot determine the correct answer, make your best guess.\\
You must use \$\$...\$\$ to enclose the label of your final answer. For example, if your final answer is K, your response should contain \$\$K\$\$.

Problem:\\
$\{$question$\}$\\

Solution:\\
$<$think$>$
\end{prompt}
\end{table*}

\begin{table*}[!h]
\caption{Generator prompt (instruction-tuned model) for MBPP+.}\label{tab:prompt-answer-instruct-mbpp}
\begin{prompt}[title={}, label=]
You are a Python programmer.\\

$\{$question$\}$
\end{prompt}
\end{table*}

\begin{table*}[!h]
\caption{Generator prompt (reasoning model) for MBPP+.}\label{tab:prompt-answer-reasoning-mbpp}
\begin{prompt}[title={}, label=]
You are a Python programmer.\\

$\{$question$\}$\\
$<$think$>$
\end{prompt}
\end{table*}

\clearpage
\section{\textcolor{black}{Additional Results on Verifiable Benchmarks}}
\subsection{\textcolor{black}{Disentangling Stylistic Bias from Objective Correctness}}

\begin{figure}[!h]
    \centering
    \includegraphics[width=\textwidth]{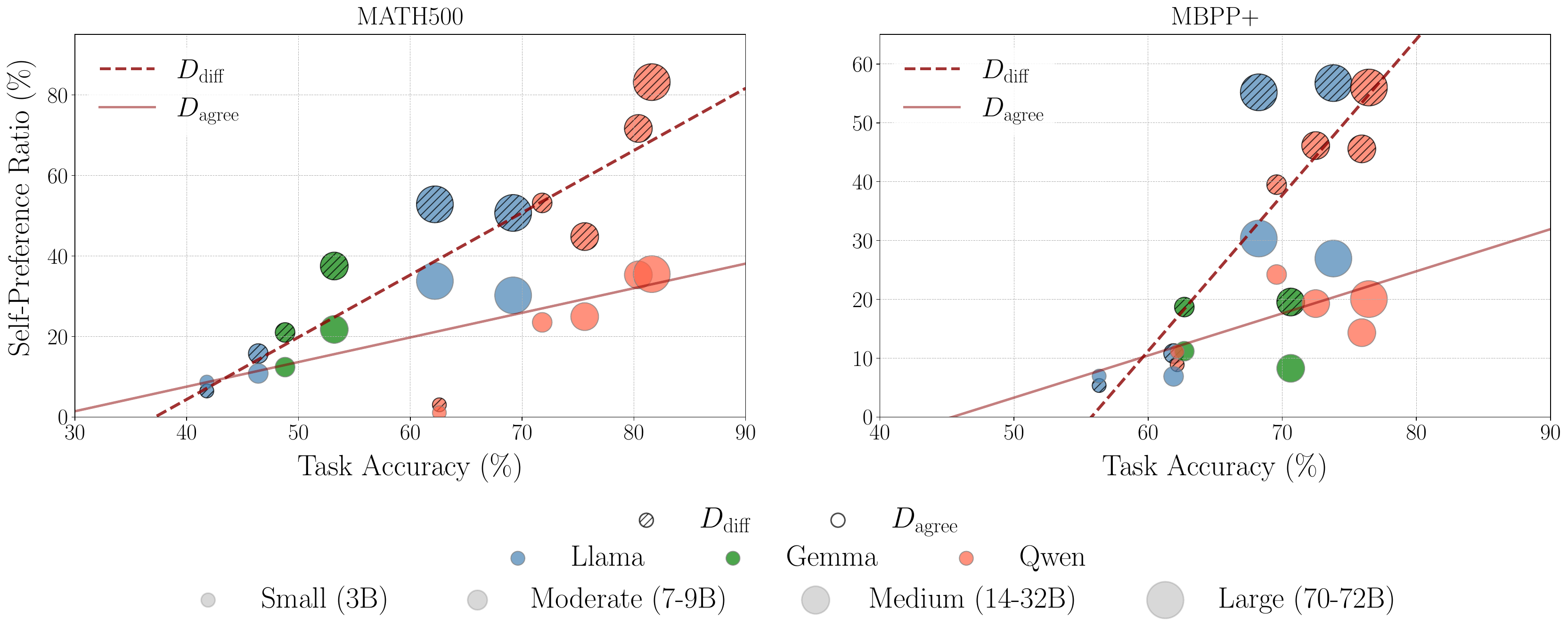} 
    \caption{\textcolor{black}{Correlation between self-preference ratio (SPR) and task accuracy.
    We decouple evaluation into correctness-driven ($\mathcal{D}_{\text{diff}}$) and style-driven ($\mathcal{D}_{\text{agree}}$) subsets. The significantly steeper slope and widening gap for $\mathcal{D}_{\text{diff}}$ indicate that as models become more capable, their self-preference is increasingly driven by the objective quality of the response rather than stylistic bias.
    We provide full results in \Cref{tab:full-res-spr-math_d_diff,tab:full-res-spr-math-d_agree,tab:full-res-spr-mmlu_d_diff,tab:full-res-spr-mmlu_d_agree,tab:full-res-spr-mbpp_d_diff,tab:full-res-spr-mbpp_d_agree}.}}
    \label{fig:d_diff_vs_d_agree}
\end{figure}

\textcolor{black}{
In \Cref{fig:d_diff_vs_d_agree}, we report SPR of various models under $\mathcal{D}_{\text{diff}}$ (instances where either $y_{\mathcal{J}}$ or $y_{\mathcal{G}}$ is correct, but not both) and $\mathcal{D}_{\text{agree}}$ (instances where both $y_{\mathcal{J}}$ and $y_{\mathcal{G}}$ are correct or incorrect).
This setup aims to disentangle the model's stylistic bias from its identification of objective correctness: 
On $\mathcal{D}_{\text{diff}}$, preference implies recognizing the correct solution; 
on $\mathcal{D}_{\text{agree}}$, the preference is driven primarily by style or formatting.
Across both mathematical reasoning (MATH500) and code generation (MBPP+), the overall trend line for $\mathcal{D}_{\text{diff}}$ is significantly steeper and reaches higher values than that of $\mathcal{D}_{\text{agree}}$.
This indicates that self-preference is not merely a result of stylistic bias.
Notably, as task accuracy increases, the gap between $\mathcal{D}_{\text{diff}}$ and $\mathcal{D}_{\text{agree}}$ widens substantially: 
High-performing models exhibit a massive spike in self-preference when answers disagree ($\mathcal{D}_{\text{diff}}$), while their preference based on style alone ($\mathcal{D}_{\text{agree}}$) grows at a much slower rate.
Coupled with the observation that legitimate self-preference ratio correlates with task performance (shown in \Cref{fig:corr_task_acc_legit_sp_ratio}), this demonstrates that as models become more capable, their self-preference is increasingly grounded in the objective correctness signal rather than stylistic familiarity.
}

\subsection{\textcolor{black}{Legitimate Self-Preference Under Capability-Matched Conditions}}

\textcolor{black}{
To rule out the possibility that high self-preference is merely an artifact of an evaluator overpowering a weaker evaluatee, we conduct a controlled experiment (reported in \Cref{tab:same-scale-math-lspr,tab:same-scale-mmlu-lspr,tab:same-scale-mbpp-lspr}) where evaluators are paired with evaluatees of comparable parametric scale and task accuracy. 
This design effectively isolates absolute model capability by neutralizing the ``performance gap'' confounder.
In this setting, a high Legitimate Self-Preference Ratio (LSPR) cannot be attributed to the evaluator simply being predominantly superior to the comparison model.
The results reveal a consistent positive correlation between LSPR and model capability: larger, high-performing models possess a more notable ``truth-detection'' capability that allows them to confidently validate their own correct answers, even when the alternative is generated by an equally powerful model.
}

\begin{table}[htbp]
  \small
  \centering
  \caption{\color{black}Legitimate self-preference ratio under evaluator-evaluatee pairs with similar parametric scale and task accuracy on MATH500. Stronger evaluators consistently record higher LSPR, reinforcing our findings in \Cref{sec:legitimate} and \Cref{fig:corr_task_acc_legit_sp_ratio} by disentangling potential confounders arising from evaluator–evaluatee capability mismatches.}
    \begin{tabular}{>{\color{black}}l >{\color{black}}r >{\color{black}}r >{\color{black}}r}
    \toprule
          & \textbf{Parameter} & \textbf{Task Accuracy} & \textbf{LSPR} \\
    \midrule
    Evaluator: Phi-3.5 & 3.8B    & 44.60 & \multirow{2}[2]{*}{12.70} \\
    Evaluatee: Llama-3.2-3B & 3B  & 41.80 &  \\
    \midrule
    Evaluator: Gemma-2-27B & 27B   & 53.20 & \multirow{2}[2]{*}{73.33} \\
    Evaluatee: Mistral-Small & 22B   & 55.60 &  \\
    \midrule
    Evaluator: Llama-3.3-70B & 70B   & 69.20 & \multirow{2}[2]{*}{93.33} \\
    Evaluatee: GPT-4o & --   & 73.20 &  \\
    \bottomrule
    \end{tabular}
  \label{tab:same-scale-math-lspr}
\end{table}

\begin{table}[htbp]
  \small
  \centering
  \caption{\color{black}Legitimate self-preference ratio under evaluator-evaluatee pairs with similar parametric scale and task accuracy on MMLU. Stronger evaluators consistently record higher LSPR, reinforcing our findings in \Cref{sec:legitimate} and \Cref{fig:corr_task_acc_legit_sp_ratio} by disentangling potential confounders arising from evaluator–evaluatee capability mismatches.}
    \begin{tabular}{>{\color{black}}l >{\color{black}}r >{\color{black}}r >{\color{black}}r}
    \toprule
          & \textbf{Parameter} & \textbf{Task Accuracy} & \textbf{LSPR} \\
    \midrule
    Evaluator: Llama-3.2-3B & 3B    & 41.80 & \multirow{2}[2]{*}{45.45} \\
    Evaluatee: Gemma-2-2B & 2B  & 42.90 &  \\
    \midrule
    Evaluator: Gemma-2-27B & 27B   & 73.80 & \multirow{2}[2]{*}{53.57} \\
    Evaluatee: Mistral-Small & 22B   & 76.10 &  \\
    \midrule
    Evaluator: Llama-3.3-70B & 70B   & 83.20 & \multirow{2}[2]{*}{68.75} \\
    Evaluatee: GPT-4o & --   & 86.60 &  \\
    \bottomrule
    \end{tabular}
  \label{tab:same-scale-mmlu-lspr}
\end{table}

\begin{table}[htbp]
  \small
  \centering
  \caption{\color{black}Legitimate self-preference ratio under evaluator-evaluatee pairs with similar parametric scale and task accuracy on MBPP+. Stronger evaluators consistently record higher LSPR, reinforcing our findings in \Cref{sec:legitimate} and \Cref{fig:corr_task_acc_legit_sp_ratio} by disentangling potential confounders arising from evaluator–evaluatee capability mismatches.}
    \begin{tabular}{>{\color{black}}l >{\color{black}}r >{\color{black}}r >{\color{black}}r}
    \toprule
          & \textbf{Parameter} & \textbf{Task Accuracy} & \textbf{LSPR} \\
    \midrule
    Evaluator: Qwen2.5-1.5B & 1.5B    & 50.79 & \multirow{2}[2]{*}{25.00} \\
    Evaluatee: Gemma-2-2B & 2B    & 49.21 &  \\
    \midrule
    Evaluator: Qwen2.5-3B & 3B   & 62.17 & \multirow{2}[2]{*}{33.33} \\
    Evaluatee: Phi-3.5 & 3.8B   & 62.43 &  \\
    \midrule
    Evaluator: Llama-3.3-70B & 70B   & 73.81 & \multirow{2}[2]{*}{73.69} \\
    Evaluatee: GPT-3.5-Turbo & --   & 70.37 &  \\
    \bottomrule
    \end{tabular}
  \label{tab:same-scale-mbpp-lspr}
\end{table}

\newpage
\subsection{\textcolor{black}{Effect of Different Decoding Configurations on Long CoT Mitigation.}}
\textcolor{black}{To evaluate the robustness of long-CoT mitigation for harmful self-preference propensity under different decoding configurations, we experiment with different decoding temperatures $\{0.4, 0.6, 0.8\}$ on MATH500 and report results in \Cref{tab:decoding}. As shown, long CoT reasoning consistently achieves lower HSPP scores across all tested temperatures, demonstrating that its mitigating effect is stable under varying decoding settings.
We adopt Llama-3.1-8B for the no-reasoning baseline and DeepSeek-R1-Distill-Llama-8B for the long CoT setting, and report statistical significance following recommended practices by~\citet{dror-etal-2018-hitchhikers}.\footnote{\url{https://github.com/rtmdrr/testSignificanceNLP}}}

\begin{table}[htbp]
\centering
\small
\caption{\color{black}Harmful self-preference propensity on MATH500 under different decoding temperatures for long CoT evaluators. Long CoT methods remain robust across all decoding temperatures and consistently achieve statistically significant reductions in HSPP. We apply McNemar’s test~\citep{mcnemar1947note}, where $\dag$ indicates $p < 0.01$.}
\label{tab:addlabel}

\begin{tabular}{>{\color{black}}l >{\color{black}}r}
\toprule
\textbf{Method} & \textbf{HSPP} \\
\midrule
No reasoning & 29.53 \\
\midrule
\multicolumn{2}{>{\color{black}}l}{Long CoT reasoning} \\
\quad - \textit{Temperature = 0.4} & 11.87\textsuperscript{\dag} \\
\quad - \textit{Temperature = 0.6} & 15.34\textsuperscript{\dag} \\
\quad - \textit{Temperature = 0.8} & 15.51\textsuperscript{\dag} \\
\bottomrule
\end{tabular}

\label{tab:decoding}
\end{table}

\subsection{\textcolor{black}{Judgment Confidence in (Harmful) Self-Preference Instances}}
\color{black}
To explore whether the internal signals of evaluators might correlate with potentially harmful self-preference judgments, we examine the average verdict confidence (\ie, probability) associated with the judgment $\mathcal{J}^* = j_1$ (preferring model $\mathcal{A}$ itself),
under the SPR set (instances which receive a final verdict $\mathcal{J}^* = j_1$) and the HSPP set (instances which receive a final verdict $\mathcal{J}^* = j_1$ while being objectively incorrect).
As shown in \Cref{tab:log_prob}, compared to the SPR set, the HSPP set consistently exhibits a lower confidence value associated with making the judgment $\mathcal{J}^* = j_1$.
This demonstrates that models might exhibit internal doubts when incorrectly favoring their own outputs.
Notably, this confidence gap between SPR and HSPP correlates strongly with model capability:
as models scale, they become better calibrated. 
This latent uncertainty signal in larger models offers a promising avenue for automatically detecting and mitigating harmful self-preference.

\begin{table}[htbp]
  \centering
  \small
  \caption{\color{black}The judgment confidence of self-preferred instances in full set and harmful set.}
    \begin{tabular}{>{\color{black}}l >{\color{black}}l >{\color{black}}r}
    \toprule
          &   \multicolumn{1}{>{\color{black}}l}{\textbf{Set}}   & \multicolumn{1}{>{\color{black}}l}{\textbf{\shortstack{Avg. probability of $\mathcal{J}^* = j_1$ \\ (Prefer itself)}}} \\
    \midrule
    \multirow{2}[2]{*}{Llama-3.2-3B} 
          & SPR   & 31.14 \\
          & HSPP  & 30.60 \\
    \midrule
    \multirow{2}[2]{*}{Llama-3.1-8B} 
          & SPR   & 57.01 \\
          & HSPP  & 48.81 \\
    \midrule
    \multirow{2}[2]{*}{Llama-3.1-70B} 
          & SPR   & 83.83 \\
          & HSPP  & 64.99 \\
    \bottomrule
    \end{tabular}%
  \label{tab:log_prob}%
\end{table}%

\clearpage
\section{\textcolor{black}{Dataset details of Subjective Tasks}}\label{app:sub-data}
\textcolor{black}{We apply the following preprocessing on the LMArena dataset for our subjective task experiments:}
\paragraph{\textcolor{black}{Models.}}
\textcolor{black}{
For consistency, we select samples in which at least one of the responses is generated by an identical or similar family model to those used as evaluators in our verifiable benchmark experiments (\Cref{sec:models})
\paragraph{Language.}
The original dataset contains user queries and model responses in multiple languages.
We further filter and retain only the samples in English.
\paragraph{Conversation turns.}
Most samples are single-turn conversations; we remove the pairs that contain multi-turn conversations during filtering.
\paragraph{Prompt length.}
To ensure the context window of evaluators are sufficient, we filter out samples in which the combined token length of the question and the responses from model $\mathcal{A}$ and model $\mathcal{B}$ exceeds 4096.
\paragraph{Response refusal.}
We filter out samples where either model refuses to provide an answer in their responses to enable a more fair and direct comparison for evaluators.}

\newpage
\section{\textcolor{black}{Evaluatee Distributions of Subjective Tasks}\label{app:sub-evaluatee-distri}}
\begin{figure}[!h]
    \centering
    \includegraphics[width=0.72\textwidth]{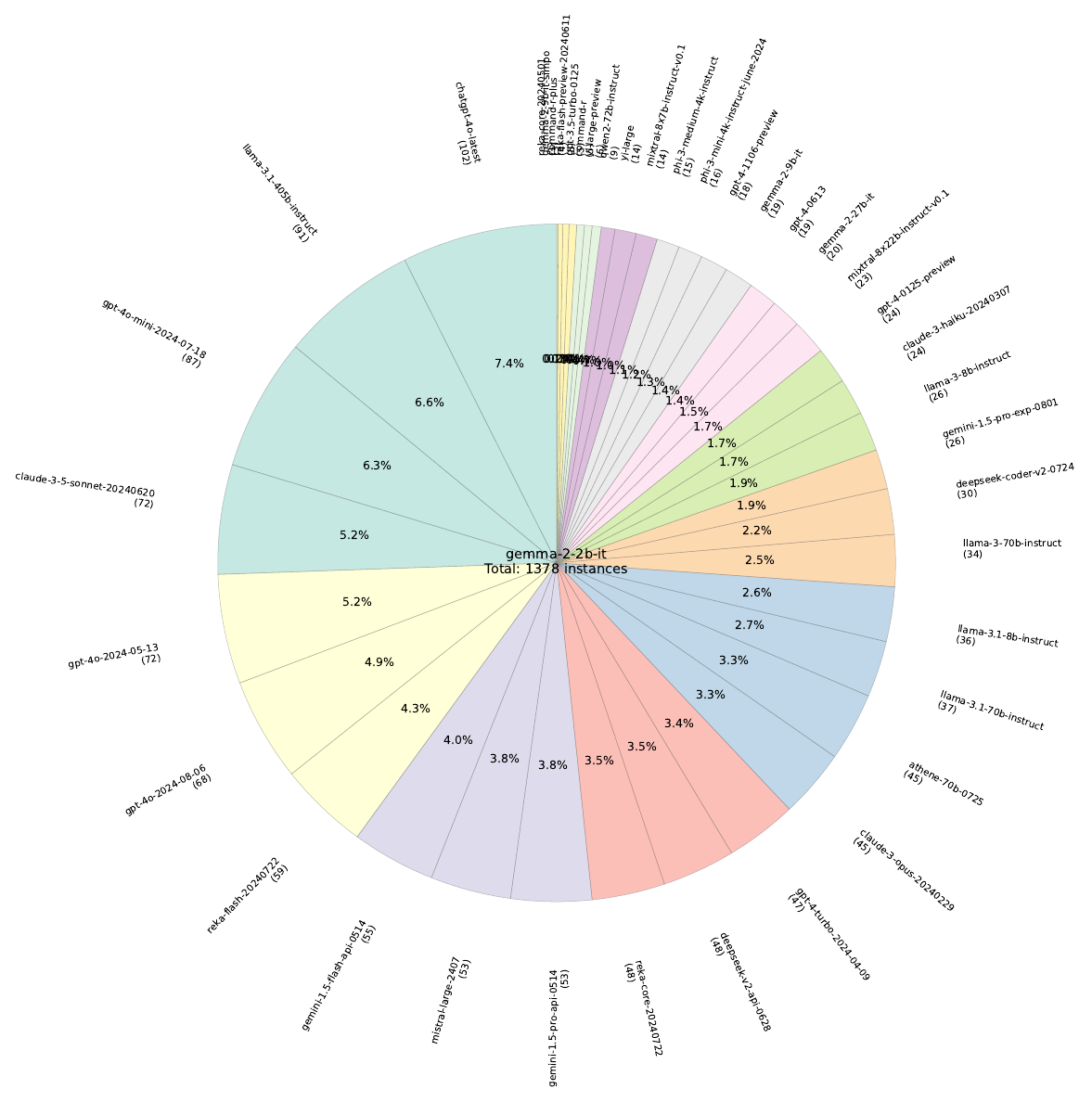} 
    \caption{\textcolor{black}{The distribution of evaluatees when evaluator is Gemma-2-2B.}}
    \label{fig:sub-evaluatee-gemma-2-2b}
    \vspace{-1em}
\end{figure}

\begin{figure}[!h]
    \centering
    \includegraphics[width=0.72\textwidth]{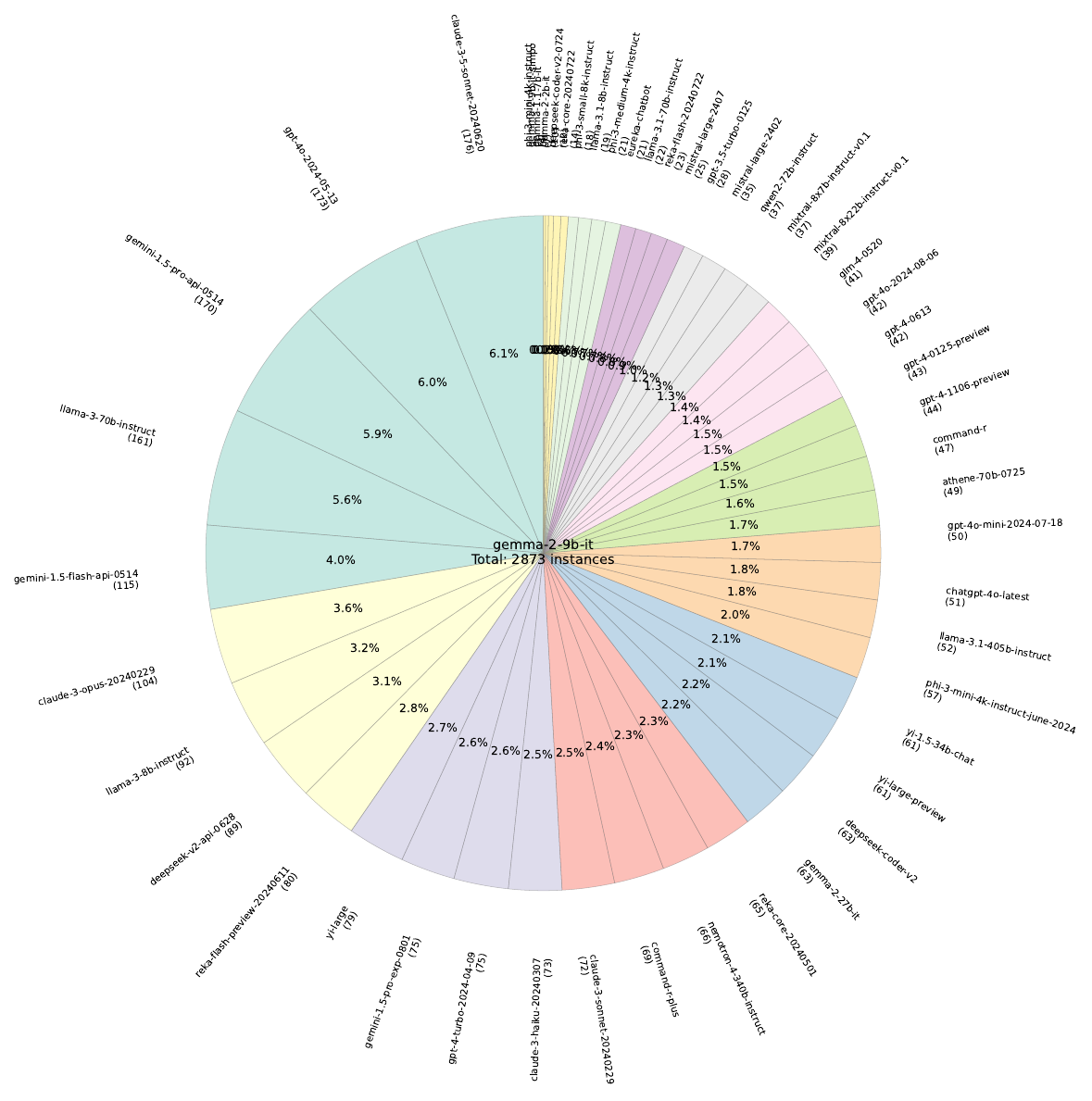} 
    \caption{\textcolor{black}{The distribution of evaluatees when evaluator is Gemma-2-9B.}}
    \label{fig:sub-evaluatee-gemma-2-9b}
    \vspace{-1em}
\end{figure}

\begin{figure}[!t]
    \centering
    \includegraphics[width=0.725\textwidth]{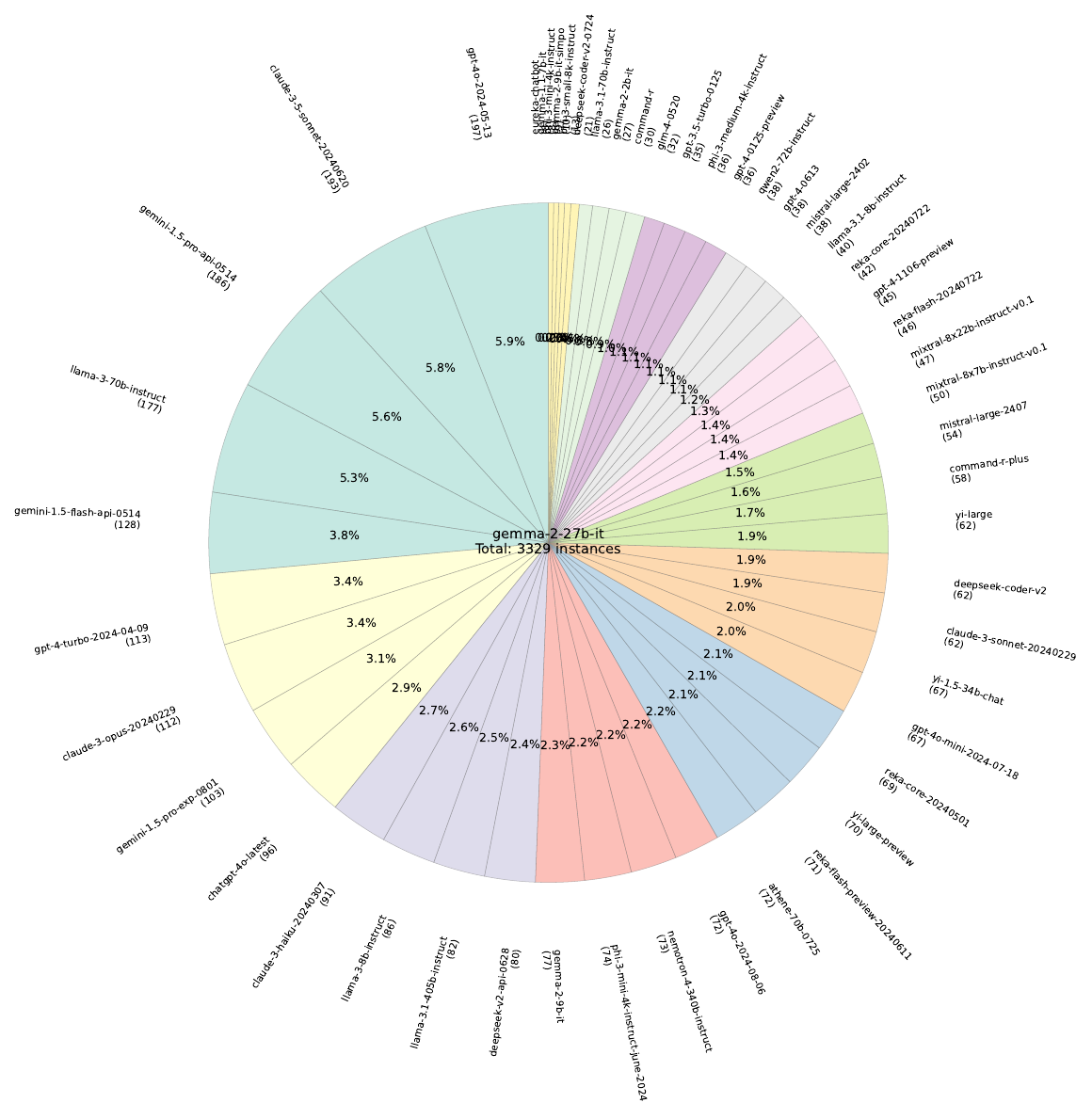} 
    \caption{\textcolor{black}{The distribution of evaluatees when evaluator is Gemma-2-27B.}}
    \label{fig:sub-evaluatee-gemma-2-27b}
    \vspace{-1em}
\end{figure}

\begin{figure}[!t]
    \centering
    \includegraphics[width=0.725\textwidth]{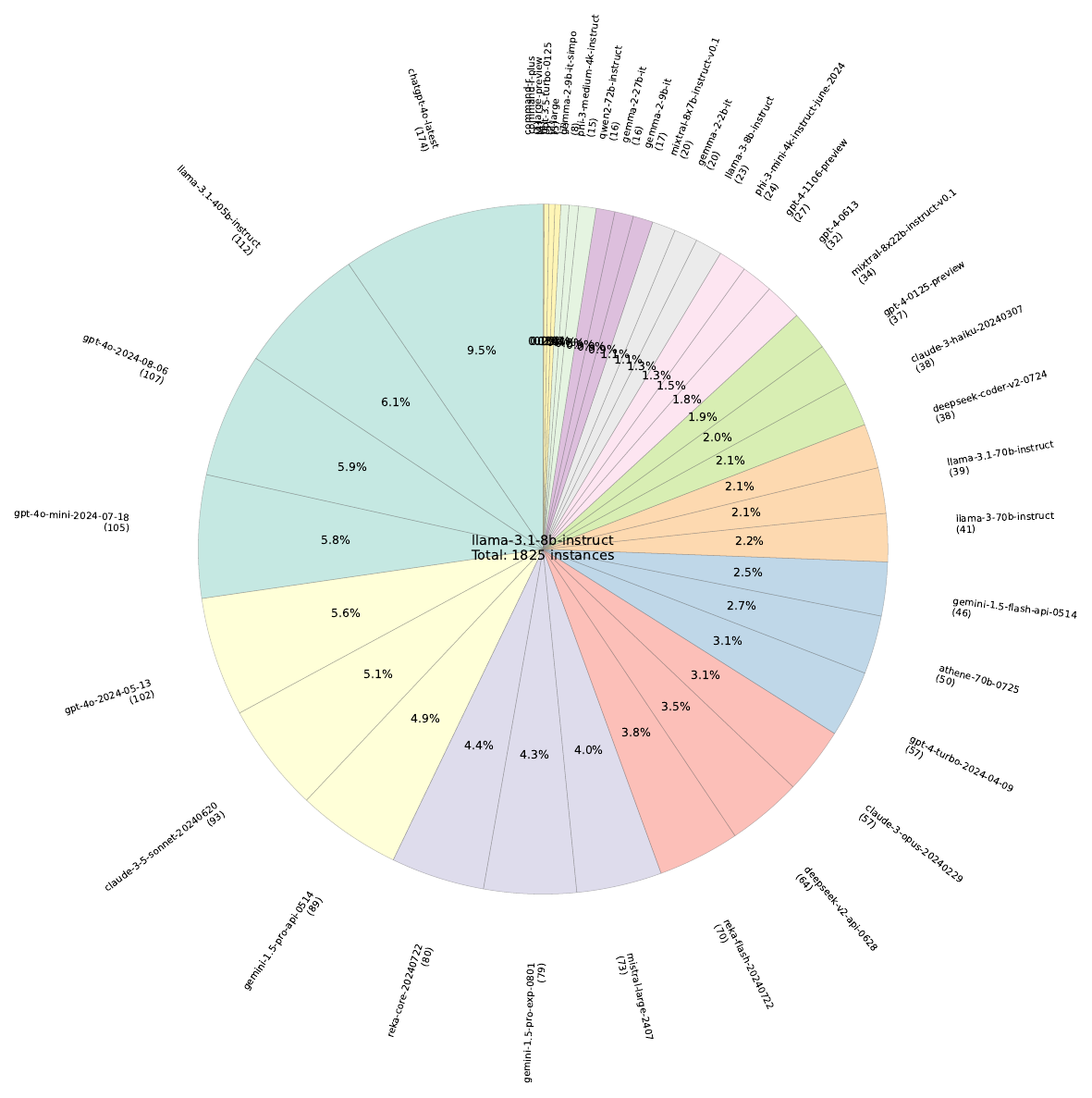} 
    \caption{\textcolor{black}{The distribution of evaluatees when evaluator is Llama-3.1-8B.}}
    \label{fig:sub-evaluatee-llama-3.1-8b}
    \vspace{-1em}
\end{figure}

\begin{figure}[!t]
    \centering
    \includegraphics[width=0.725\textwidth]{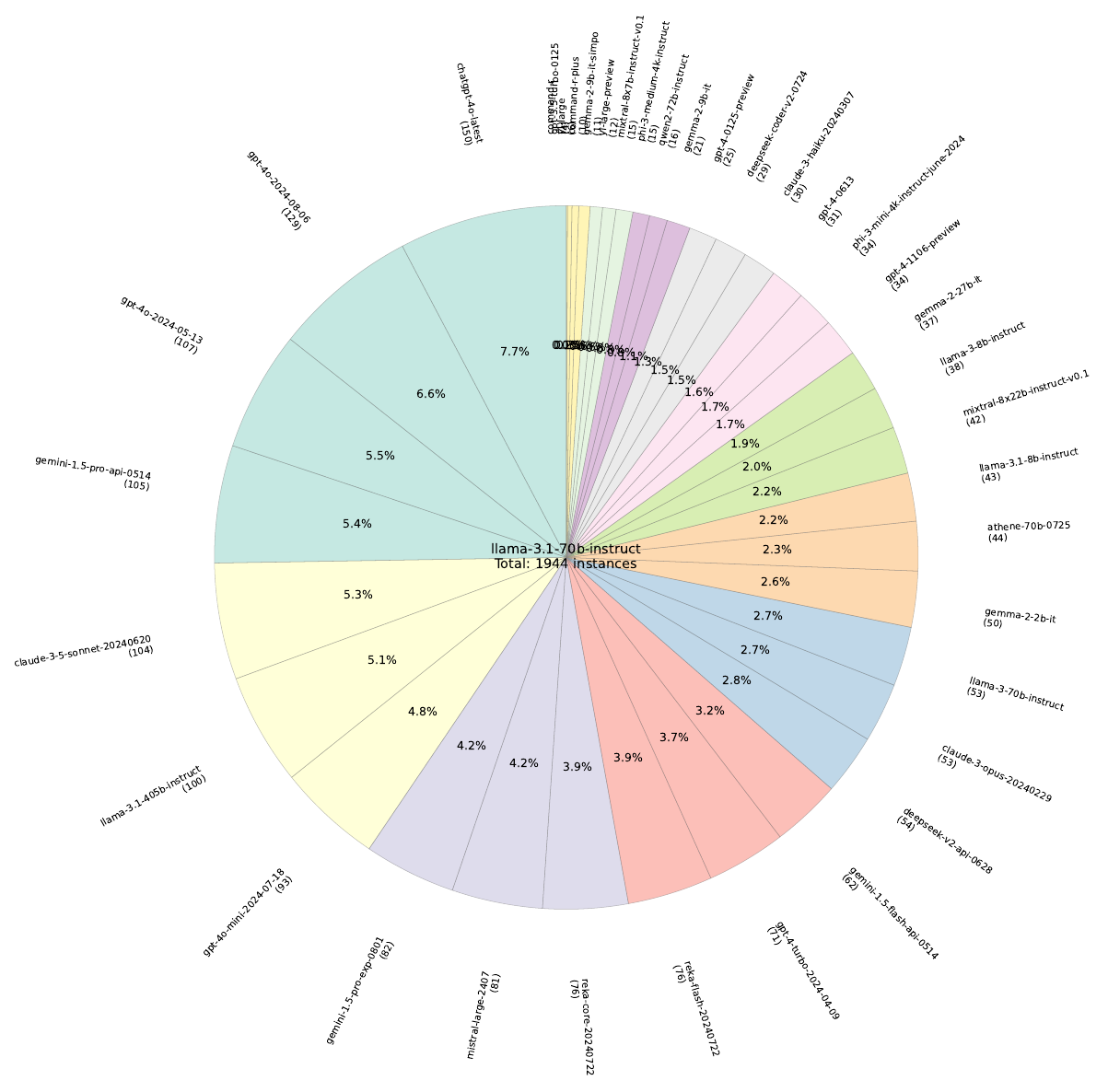} 
    \caption{\textcolor{black}{The distribution of evaluatees when evaluator is Llama-3.1-70B.}}
    \label{fig:sub-evaluatee-llama-3.1-70b}
    \vspace{-1em}
\end{figure}

\begin{figure}[!t]
    \centering
    \includegraphics[width=0.725\textwidth]{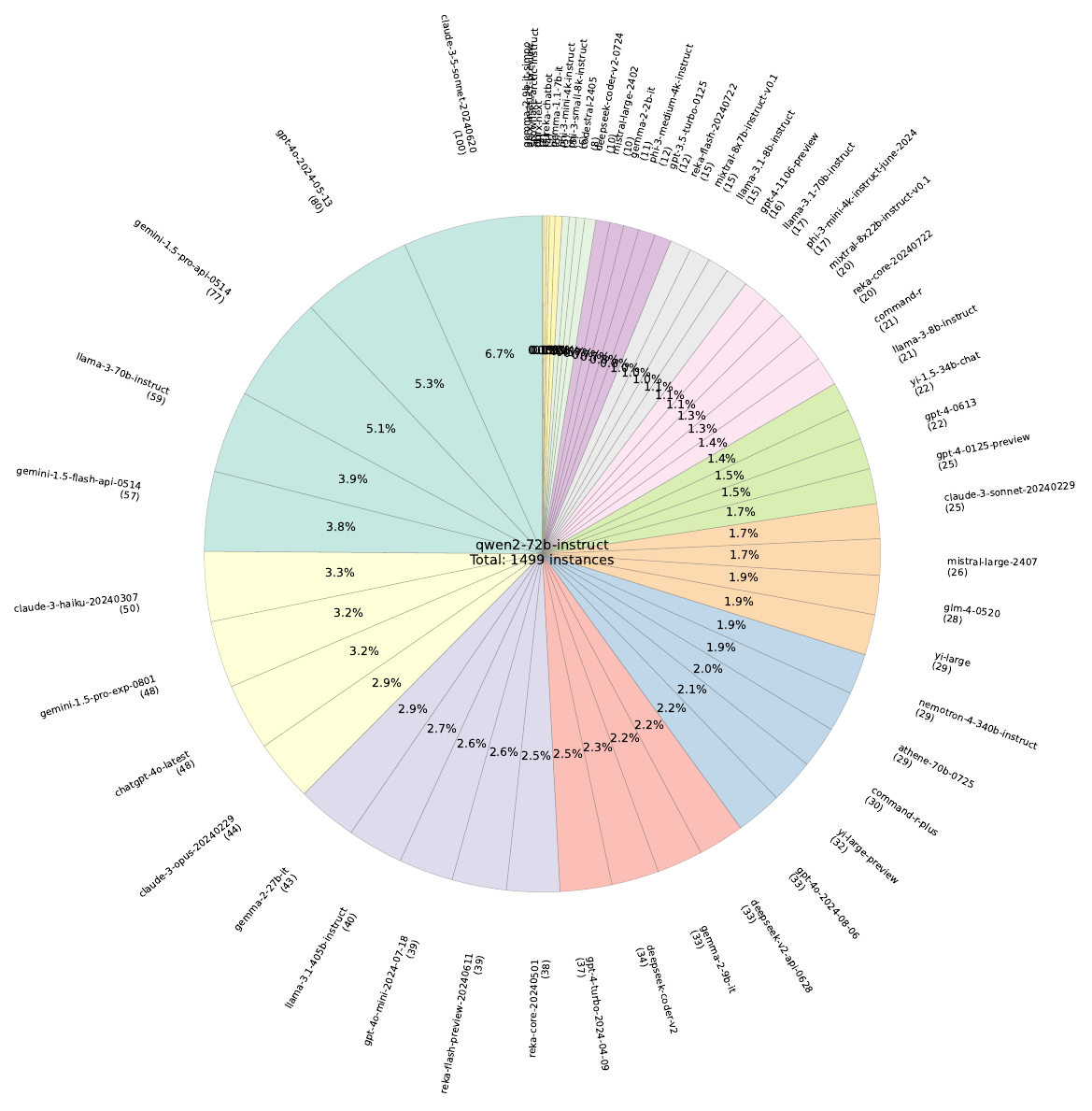} 
    \caption{\textcolor{black}{The distribution of evaluatees when evaluator is Qwen2-72B.}}
    \label{fig:sub-evaluatee-qwen2-72b}
    \vspace{-1em}
\end{figure}

We provide the corresponding evaluatee distributions for each evaluator in \Cref{fig:sub-evaluatee-gemma-2-2b,fig:sub-evaluatee-gemma-2-9b,fig:sub-evaluatee-gemma-2-27b,fig:sub-evaluatee-llama-3.1-8b,fig:sub-evaluatee-llama-3.1-70b,fig:sub-evaluatee-qwen2-72b}, including the total number of samples for each evaluator and a fine-grained breakdown of the paired evaluatee models.

\clearpage
\newpage

{\color{black}
\section{Case Study on Code Generation Task}
\label{app:case-study}
We present a case study on the coding task MBPP+, demonstrating that performant models that produce correct code accurately identify bugs and errors in peer code and produce a legitimate preference verdict aligning with objective quality of responses.

\Cref{tab:case-70b-answers} and \Cref{tab:case-gpt-answers} present the responses generated by a similarly capable model pair---Llama-3.1-70B (Model $\mathcal{A}$) and GPT-3.5-Turbo (Model $\mathcal{B}$)---to the same code-generation question in \Cref{tab:case-question}, where $\mathcal{A}$ produces a correct solution and $\mathcal{B}$ produces a incorrect one.
\Cref{tab:case-70b-judge} shows the judgment produced by the evaluator model (Model $\mathcal{A}$), Llama-3.1-70B. As shown, Llama-3.1-70B correctly identifies the error in GPT-3.5-Turbo’s answer regarding the misuse of the Python module with incorrect index subtraction (highlighted in yellow in \Cref{tab:case-70b-judge}), and therefore issues a legitimate self-preference verdict in favor of $\mathcal{A}$.}

\begin{table*}[!h]
\caption{\textcolor{black}{A code generation question from MBPP+.}}\label{tab:case-question}
\begin{prompt}[title={}, label=]
\color{black}Write a function to search a string for a regex pattern. The function should return the matching subtring, a start index and an end index. Do not include other descriptions except comments for the code.\\

Your code should satisfy these tests:\\
\texttt{
assert find\_literals('The quick brown fox jumps over the lazy dog.', 'fox') == ('fox', 16, 19)\\
assert find\_literals('Its been a very crazy procedure right', 'crazy') == ('crazy', 16, 21)\\
assert find\_literals('Hardest choices required strongest will', 'will') == ('will', 35, 39)
}
\end{prompt}
\end{table*}

\begin{table*}[!h]
\caption{\textcolor{black}{Llama-3.1-70B (Model $\mathcal{A}$)'s \textcolor{green}{correct} response to the question.}}\label{tab:case-70b-answers}
\begin{prompt}[title={}, label=]
\color{black}
\texttt{import re\\
def find\_literals(string, pattern):\\
\hspace*{2em}    \# Search the string for the pattern\\
\hspace*{2em}    match = re.search(pattern, string)\\
\hspace*{2em}    \# If a match is found, return the matched substring and its start\\
\hspace*{2.5em}and end indices\\
\hspace*{2em}    if match:\\
\hspace*{4em}        return (match.group(), match.start(), match.end())\\
\hspace*{2em}    \# If no match is found, return None\\
\hspace*{2em}    else:\\
\hspace*{4em}        return None
}
\end{prompt}
\end{table*}

\begin{table*}[!h]
\caption{\textcolor{black}{GPT-3.5-Turbo (Model $\mathcal{B}$)'s \textcolor{red}{incorrect} response to the question.}}\label{tab:case-gpt-answers}
\begin{prompt}[title={}, label=]
\color{black}
\texttt{import re\\
def find\_literals(text, pattern):\\
\hspace*{2em}    \# Search for the pattern in the text\\
\hspace*{2em}    match = re.search(pattern, text)\\
\hspace*{2em}    if match:\\
\hspace*{4em}        \# Return the matching substring, start index, and end index\\
\hspace*{4em}        return match.group(), match.start(), match.end() - 1\\
\hspace*{2em}    else:\\
\hspace*{4em}        return None}
\end{prompt}
\end{table*}

\begin{table*}[!h]
\caption{\textcolor{black}{Llama-3.1-70B's judgments to the response pair of Model $\mathcal{A}$ and Model $\mathcal{B}$.}}\label{tab:case-70b-judge}
\begin{prompt}[title={}, label=]
\color{black}
After reviewing the provided functions, I noticed that both functions are almost identical and use the `re.search` function from Python's built-in `re` module to search for the pattern in the string. However, there is a subtle difference between the two functions.\\

Assistant A's function returns the end index of the match as is, \colorbox{yellow}{whereas Assistant B's function returns the} \colorbox{yellow}{end index minus 1. The `end()` method in Python's `re` module returns the ending position of the match,} \colorbox{yellow}{which is the index after the last character of the match. Therefore, if you want to get the index of the last} \colorbox{yellow}{character of the match, you should not subtract 1 from the end index.}\\

Considering the test cases provided, Assistant A's function returns the correct end indices, whereas Assistant B's function returns incorrect end indices.\\
My final verdict is \$\$A\$\$.
\end{prompt}
\end{table*}

\clearpage
\color{black}{
\section{Full Experimental Results}
}
\label{app:fall-res}

\begin{table}[htbp]
  \small
  \centering
  \caption{Full SPR results of MATH500. The numbers in \Cref{fig:corre_task_acc_sp_ratio} correspond to the average column.}
  \begin{tabular}{lrrrrrrr>{\columncolor{gray!20}}r}
    \toprule
    & \multicolumn{8}{c}{\textbf{Evaluatee}} \\
    \cmidrule{2-9}
    \textbf{Evaluator} & \multicolumn{1}{c}{\rotatebox{90}{Mistral-7b}} & \multicolumn{1}{c}{\rotatebox{90}{Phi-3.5}} & \multicolumn{1}{c}{\rotatebox{90}{GPT-3.5-Turbo}} & \multicolumn{1}{c}{\rotatebox{90}{GPT-4o}} & \multicolumn{1}{c}{\rotatebox{90}{Gemma-2-2B}} & \multicolumn{1}{c}{\rotatebox{90}{Llama-3.2-1B}} & \multicolumn{1}{c}{\rotatebox{90}{Mistral-Small}} & \multicolumn{1}{c}{\rotatebox{90}{Average}} \\
    \midrule
    Llama-3.2-3B & 4.2 & 5.8 & 1.0 & 12.0 & 11.0 & 17.2 & 3.8 & \cellcolor{gray!20}7.9 \\
    Llama-3.1-8B & 20.6 & 12.8 & 7.2 & 5.2 & 22.8 & 11.6 & 6.6 & \cellcolor{gray!20}12.4 \\
    Llama-3.1-70B & 68.4 & 36.4 & 34.6 & 12.6 & 59.0 & 56.4 & 16.8 & \cellcolor{gray!20}40.6 \\
    Llama-3.3-70B & 66.4 & 34.6 & 31.8 & 9.2 & 56.8 & 51.0 & 18.0 & \cellcolor{gray!20}38.3 \\
    \midrule
    Gemma-2-9B & 25.2 & 9.0 & 11.4 & 6.2 & 28.2 & 19.8 & 5.2 & \cellcolor{gray!20}15.0 \\
    Gemma-2-27B & 39.6 & 20.8 & 20.4 & 8.2 & 45.4 & 38.6 & 12.8 & \cellcolor{gray!20}26.5 \\
    \midrule
    Qwen2.5-3B & 1.6 & 1.2 & 2.4 & 1.2 & 2.0 & 1.2 & 2.2 & \cellcolor{gray!20}1.7 \\
    Qwen2.5-7B & 61.4 & 32.4 & 26.4 & 6.8 & 54.2 & 53.8 & 13.4 & \cellcolor{gray!20}35.5 \\
    Qwen2.5-14B & 53.4 & 27.6 & 28.6 & 8.4 & 55.6 & 33.6 & 16.2 & \cellcolor{gray!20}31.9 \\
    Qwen2.5-32B & 82.6 & 43.8 & 51.4 & 11.8 & 70.8 & 71.2 & 27.2 & \cellcolor{gray!20}51.3 \\
    Qwen2.5-72B & 86.4 & 50.6 & 56.4 & 15.8 & 72.6 & 73.2 & 31.2 & \cellcolor{gray!20}55.2 \\
    \bottomrule
  \end{tabular}
  \label{tab:full-res-spr-math}
\end{table}

\begin{table}[htbp]
  \small
  \centering
  \caption{Full HSPP results of MATH500. The numbers in \Cref{fig:corre_task_acc_harmful_sp_ratio} correspond to the average column.}
  \begin{tabular}{lrrrrrrr>{\columncolor{gray!20}}r}
    \toprule
    & \multicolumn{8}{c}{\textbf{Evaluatee}} \\
    \cmidrule{2-9}
    \textbf{Evaluator} & \multicolumn{1}{c}{\rotatebox{90}{Mistral-7b}} & \multicolumn{1}{c}{\rotatebox{90}{Phi-3.5}} & \multicolumn{1}{c}{\rotatebox{90}{GPT-3.5-Turbo}} & \multicolumn{1}{c}{\rotatebox{90}{GPT-4o}} & \multicolumn{1}{c}{\rotatebox{90}{Gemma-2-2B}} & \multicolumn{1}{c}{\rotatebox{90}{Llama-3.2-1B}} & \multicolumn{1}{c}{\rotatebox{90}{Mistral-Small}} & \multicolumn{1}{c}{\rotatebox{90}{Average}} \\
    \midrule
    Llama-3.2-3B & 38.1 & 34.4 & 25.4 & 29.2 & 20.0 & 52.2 & 29.6 & \cellcolor{gray!20}32.7 \\
    Llama-3.1-8B & 43.5 & 32.1 & 22.8 & 19.0 & 43.5 & 29.2 & 16.7 & \cellcolor{gray!20}29.5 \\
    Llama-3.1-70B & 50.0 & 25.0 & 34.8 & 20.9 & 44.4 & 75.0 & 25.6 & \cellcolor{gray!20}39.4 \\
    Llama-3.3-70B & 75.0 & 20.0 & 26.7 & 20.0 & 28.6 & 41.7 & 28.0 & \cellcolor{gray!20}34.3 \\
    \midrule
    Gemma-2-9B & 69.2 & 53.5 & 53.7 & 43.9 & 42.9 & 90.0 & 42.7 & \cellcolor{gray!20}56.5 \\
    Gemma-2-27B & 81.8 & 58.3 & 50.0 & 42.1 & 77.8 & 81.0 & 54.1 & \cellcolor{gray!20}63.6 \\
    \midrule
    Qwen2.5-3B & 71.4 & 57.9 & 55.0 & 36.3 & 71.4 & 72.7 & 37.8 & \cellcolor{gray!20}57.5 \\
    Qwen2.5-7B & 66.7 & 72.7 & 27.3 & 51.3 & 80.0 & 62.5 & 38.1 & \cellcolor{gray!20}56.9 \\
    Qwen2.5-14B & 80.0 & 83.3 & 71.4 & 23.3 & 100.0 & 16.7 & 41.7 & \cellcolor{gray!20}59.5 \\
    Qwen2.5-32B & 100.0 & 100.0 & 80.0 & 33.3 & 75.0 & 71.4 & 37.5 & \cellcolor{gray!20}71.0 \\
    Qwen2.5-72B & 100.0 & 100.0 & 100.0 & 53.3 & 100.0 & 100.0 & 50.0 & \cellcolor{gray!20}86.2 \\
    \bottomrule
  \end{tabular}
  \label{tab:full-res-hspr-math}
\end{table}

\begin{table}[htbp]
  \small
  \centering
  \caption{Full SPR results of MMLU. The numbers in \Cref{fig:corre_task_acc_sp_ratio} correspond to the average column.}  
  \begin{tabular}{lrrrrrrr>{\columncolor{gray!20}}r}
    \toprule
    & \multicolumn{8}{c}{\textbf{Evaluatee}} \\
    \cmidrule{2-9}
    \textbf{Evaluator} & \multicolumn{1}{c}{\rotatebox{90}{Mistral-7b}} & \multicolumn{1}{c}{\rotatebox{90}{Phi-3.5}} & \multicolumn{1}{c}{\rotatebox{90}{GPT-3.5-Turbo}} & \multicolumn{1}{c}{\rotatebox{90}{GPT-4o}} & \multicolumn{1}{c}{\rotatebox{90}{Gemma-2-2B}} & \multicolumn{1}{c}{\rotatebox{90}{Llama-3.2-1B}} & \multicolumn{1}{c}{\rotatebox{90}{Mistral-Small}} & \multicolumn{1}{c}{\rotatebox{90}{Average}} \\
    \midrule
    Llama-3.2-3B & 2.2 & 4.3 & 4.2 & 8.3 & 4.5 & 8.0 & 3.6 & \cellcolor{gray!20}5.0 \\
    Llama-3.1-8B & 37.2 & 11.9 & 19.7 & 2.6 & 18.2 & 52.8 & 17.6 & \cellcolor{gray!20}22.9 \\
    Llama-3.1-70B & 63.5 & 22.4 & 37.5 & 2.8 & 39.5 & 80.6 & 27.1 & \cellcolor{gray!20}39.1 \\
    Llama-3.3-70B & 74.7 & 50.8 & 64.8 & 12.9 & 57.2 & 81.5 & 60.3 & \cellcolor{gray!20}57.5 \\
    \midrule
    Gemma-2-9B & 26.1 & 8.7 & 13.9 & 2.0 & 12.2 & 44.1 & 10.4 & \cellcolor{gray!20}16.8 \\
    Gemma-2-27B & 24.4 & 8.8 & 16.1 & 5.9 & 13.5 & 49.0 & 11.7 & \cellcolor{gray!20}18.5 \\
    \midrule
    Qwen2.5-3B & 6.9 & 5.6 & 6.0 & 2.8 & 4.7 & 6.0 & 5.5 & \cellcolor{gray!20}5.4 \\
    Qwen2.5-7B & 46.9 & 11.6 & 15.8 & 2.6 & 27.9 & 76.9 & 21.7 & \cellcolor{gray!20}29.1 \\
    Qwen2.5-14B & 63.6 & 24.2 & 39.0 & 5.1 & 38.1 & 75.9 & 34.6 & \cellcolor{gray!20}40.1 \\
    Qwen2.5-32B & 84.9 & 57.8 & 66.9 & 9.2 & 61.6 & 88.4 & 66.7 & \cellcolor{gray!20}62.2 \\
    Qwen2.5-72B & 79.8 & 35.4 & 56.1 & 4.3 & 48.2 & 92.9 & 51.1 & \cellcolor{gray!20}52.5 \\
    \bottomrule
  \end{tabular}
  \label{tab:full-res-spr-mmlu}
\end{table}

\begin{table}[htbp]
  \small
  \centering
  \caption{Full HSPP results of MMLU. The numbers in \Cref{fig:corre_task_acc_harmful_sp_ratio} correspond to the average column.}
  \begin{tabular}{lrrrrrrr>{\columncolor{gray!20}}r}
    \toprule
    & \multicolumn{8}{c}{\textbf{Evaluatee}} \\
    \cmidrule{2-9}
    \textbf{Evaluator} & \multicolumn{1}{c}{\rotatebox{90}{Mistral-7b}} & \multicolumn{1}{c}{\rotatebox{90}{Phi-3.5}} & \multicolumn{1}{c}{\rotatebox{90}{GPT-3.5-Turbo}} & \multicolumn{1}{c}{\rotatebox{90}{GPT-4o}} & \multicolumn{1}{c}{\rotatebox{90}{Gemma-2-2B}} & \multicolumn{1}{c}{\rotatebox{90}{Llama-3.2-1B}} & \multicolumn{1}{c}{\rotatebox{90}{Mistral-Small}} & \multicolumn{1}{c}{\rotatebox{90}{Average}} \\
    \midrule
    Llama-3.2-3B & 21.5 & 19.5 & 26.4 & 37.8 & 24.6 & 59.9 & 21.3 & \cellcolor{gray!20}30.1 \\
    Llama-3.1-8B & 66.3 & 39.3 & 54.3 & 12.0 & 34.5 & 88.0 & 42.6 & \cellcolor{gray!20}48.1 \\
    Llama-3.1-70B & 77.6 & 41.7 & 48.8 & 10.9 & 52.5 & 90.0 & 44.2 & \cellcolor{gray!20}52.2 \\
    Llama-3.3-70B & 89.7 & 88.9 & 77.8 & 25.0 & 72.4 & 94.3 & 66.0 & \cellcolor{gray!20}73.4 \\
    \midrule
    Gemma-2-9B & 56.9 & 36.4 & 42.9 & 26.1 & 44.8 & 88.3 & 38.1 & \cellcolor{gray!20}47.6 \\
    Gemma-2-27B & 54.9 & 39.4 & 51.8 & 23.7 & 32.1 & 87.9 & 38.5 & \cellcolor{gray!20}46.9 \\
    \midrule
    Qwen2.5-3B & 60.3 & 47.3 & 56.7 & 58.7 & 62.2 & 43.0 & 48.0 & \cellcolor{gray!20}53.7 \\
    Qwen2.5-7B & 81.4 & 55.2 & 58.4 & 13.5 & 75.0 & 98.3 & 63.7 & \cellcolor{gray!20}63.6 \\
    Qwen2.5-14B & 68.4 & 70.0 & 80.0 & 31.3 & 66.7 & 98.1 & 64.6 & \cellcolor{gray!20}68.4 \\
    Qwen2.5-32B & 89.5 & 77.8 & 94.0 & 27.2 & 73.3 & 98.0 & 72.4 & \cellcolor{gray!20}76.0 \\
    Qwen2.5-72B & 91.7 & 88.9 & 88.2 & 20.7 & 54.2 & 100.0 & 70.6 & \cellcolor{gray!20}73.5 \\
    \bottomrule
  \end{tabular}
  \label{tab:full-res-hspr-mmlu}
\end{table}

\begin{table}[htbp]
  \small
  \centering
  \caption{Full SPR results of MBPP+. The numbers in \Cref{fig:corre_task_acc_sp_ratio} correspond to the average column.}  
  \begin{tabular}{lrrrrrrr>{\columncolor{gray!20}}r}
    \toprule
    & \multicolumn{8}{c}{\textbf{Evaluatee}} \\
    \cmidrule{2-9}
    \textbf{Evaluator} & \multicolumn{1}{c}{\rotatebox{90}{Mistral-7b}} & \multicolumn{1}{c}{\rotatebox{90}{Phi-3.5}} & \multicolumn{1}{c}{\rotatebox{90}{GPT-3.5-Turbo}} & \multicolumn{1}{c}{\rotatebox{90}{GPT-4o}} & \multicolumn{1}{c}{\rotatebox{90}{Gemma-2-2B}} & \multicolumn{1}{c}{\rotatebox{90}{Llama-3.2-1B}} & \multicolumn{1}{c}{\rotatebox{90}{Mistral-Small}} & \multicolumn{1}{c}{\rotatebox{90}{Average}} \\
    \midrule
    Llama-3.2-3B & 6.3 & 8.2 & 3.7 & 6.3 & 5.8 & 7.9 & 7.4 & \cellcolor{gray!20}6.5 \\
    Llama-3.1-8B & 10.3 & 7.9 & 6.1 & 7.7 & 6.1 & 6.1 & 9.5 & \cellcolor{gray!20}7.7 \\
    Llama-3.1-70B & 48.4 & 31.0 & 27.5 & 19.0 & 52.4 & 55.6 & 25.4 & \cellcolor{gray!20}37.0 \\
    Llama-3.3-70B & 52.4 & 26.2 & 20.6 & 15.1 & 50.3 & 56.9 & 24.1 & \cellcolor{gray!20}35.1 \\
    \midrule
    Gemma-2-9B & 16.4 & 13.2 & 12.2 & 8.2 & 14.3 & 16.4 & 8.7 & \cellcolor{gray!20}12.8 \\
    Gemma-2-27B & 18.0 & 10.3 & 8.5 & 4.8 & 14.0 & 17.5 & 5.3 & \cellcolor{gray!20}11.2 \\
    \midrule
    Qwen2.5-3B & 10.8 & 6.1 & 12.7 & 9.3 & 14.8 & 10.6 & 10.1 & \cellcolor{gray!20}10.6 \\
    Qwen2.5-7B & 40.7 & 26.7 & 21.2 & 13.8 & 34.7 & 42.1 & 23.0 & \cellcolor{gray!20}28.9 \\
    Qwen2.5-14B & 40.5 & 18.0 & 10.3 & 2.1 & 38.6 & 45.5 & 14.6 & \cellcolor{gray!20}24.2 \\
    Qwen2.5-32B & 43.9 & 20.1 & 15.1 & 6.3 & 41.3 & 46.6 & 16.4 & \cellcolor{gray!20}27.1 \\
    Qwen2.5-72B & 49.2 & 24.1 & 12.7 & 6.9 & 48.7 & 52.9 & 20.1 & \cellcolor{gray!20}30.7 \\
    \bottomrule
  \end{tabular}
  \label{tab:full-res-spr-mbpp}
\end{table}

\begin{table}[htbp]
  \small
  \centering
  \caption{FullHSPP results of MBPP+. The numbers in \Cref{fig:corre_task_acc_harmful_sp_ratio} correspond to the average column.}
  \begin{tabular}{lrrrrrrr>{\columncolor{gray!20}}r}
    \toprule
    & \multicolumn{8}{c}{\textbf{Evaluatee}} \\
    \cmidrule{2-9}
    \textbf{Evaluator} & \multicolumn{1}{c}{\rotatebox{90}{Mistral-7b}} & \multicolumn{1}{c}{\rotatebox{90}{Phi-3.5}} & \multicolumn{1}{c}{\rotatebox{90}{GPT-3.5-Turbo}} & \multicolumn{1}{c}{\rotatebox{90}{GPT-4o}} & \multicolumn{1}{c}{\rotatebox{90}{Gemma-2-2B}} & \multicolumn{1}{c}{\rotatebox{90}{Llama-3.2-1B}} & \multicolumn{1}{c}{\rotatebox{90}{Mistral-Small}} & \multicolumn{1}{c}{\rotatebox{90}{Average}} \\
    \midrule
    Llama-3.2-3B & 46.7 & 55.8 & 35.6 & 48.2 & 29.4 & 52.4 & 42.9 & \cellcolor{gray!20}44.4 \\
    Llama-3.1-8B & 42.9 & 85.3 & 74.0 & 73.1 & 45.5 & 50.0 & 86.8 & \cellcolor{gray!20}65.4 \\
    Llama-3.1-70B & 66.7 & 78.6 & 71.4 & 53.5 & 72.7 & 66.7 & 64.0 & \cellcolor{gray!20}67.6 \\
    Llama-3.3-70B & 37.5 & 53.3 & 45.5 & 41.4 & 64.3 & 16.7 & 50.0 & \cellcolor{gray!20}44.1 \\
    \midrule
    Gemma-2-9B & 41.7 & 45.7 & 49.1 & 22.4 & 46.2 & 37.5 & 50.0 & \cellcolor{gray!20}41.8 \\
    Gemma-2-27B & 50.0 & 23.1 & 50.0 & 30.6 & 41.7 & 40.0 & 41.7 & \cellcolor{gray!20}39.6 \\
    \midrule
    Qwen2.5-3B & 50.0 & 34.5 & 46.0 & 58.1 & 39.3 & 46.7 & 51.5 & \cellcolor{gray!20}46.6 \\
    Qwen2.5-7B & 66.7 & 50.0 & 44.1 & 26.8 & 68.8 & 50.0 & 27.3 & \cellcolor{gray!20}47.7 \\
    Qwen2.5-14B & 55.6 & 56.3 & 46.4 & 46.4 & 61.1 & 77.8 & 31.6 & \cellcolor{gray!20}53.6 \\
    Qwen2.5-32B & 50.0 & 56.3 & 44.4 & 28.0 & 56.3 & 66.7 & 50.0 & \cellcolor{gray!20}50.2 \\
    Qwen2.5-72B & 90.0 & 53.8 & 50.0 & 57.9 & 87.5 & 50.0 & 50.0 & \cellcolor{gray!20}62.7 \\
    \bottomrule
  \end{tabular}
  \label{tab:full-res-hspr-mbpp}
\end{table}

\begin{table}[htbp]
  \small
  \centering
  \caption{Full judge accuracy results of MATH500. The numbers in \Cref{fig:corre_task_acc_judge_acc} correspond to the average column.}
  \begin{tabular}{lrrrrrrr>{\columncolor{gray!20}}r}
    \toprule
    & \multicolumn{8}{c}{\textbf{Evaluatee}} \\
    \cmidrule{2-9}
    \textbf{Evaluator} & \multicolumn{1}{c}{\rotatebox{90}{Mistral-7b}} & \multicolumn{1}{c}{\rotatebox{90}{Phi-3.5}} & \multicolumn{1}{c}{\rotatebox{90}{GPT-3.5-Turbo}} & \multicolumn{1}{c}{\rotatebox{90}{GPT-4o}} & \multicolumn{1}{c}{\rotatebox{90}{Gemma-2-2B}} & \multicolumn{1}{c}{\rotatebox{90}{Llama-3.2-1B}} & \multicolumn{1}{c}{\rotatebox{90}{Mistral-Small}} & \multicolumn{1}{c}{\rotatebox{90}{Average}} \\
    \midrule
    Llama-3.2-3B & 30.5 & 49.1 & 44.1 & 67.3 & 54.4 & 66.0 & 52.4 & \cellcolor{gray!20}52.0 \\
    Llama-3.1-8B & 70.2 & 62.8 & 54.7 & 74.4 & 68.9 & 68.1 & 69.4 & \cellcolor{gray!20}66.9 \\
    Llama-3.1-70B & 89.6 & 80.9 & 75.3 & 74.4 & 83.8 & 84.2 & 59.5 & \cellcolor{gray!20}78.2 \\
    Llama-3.3-70B & 92.0 & 85.3 & 79.9 & 75.6 & 87.1 & 91.4 & 64.4 & \cellcolor{gray!20}82.2 \\
    \midrule
    Gemma-2-9B & 80.9 & 68.2 & 64.7 & 56.4 & 66.5 & 70.6 & 60.3 & \cellcolor{gray!20}66.8 \\
    Gemma-2-27B & 83.1 & 68.7 & 65.2 & 59.2 & 78.6 & 83.5 & 61.8 & \cellcolor{gray!20}71.4 \\
    \midrule
    Qwen2.5-3B & 48.8 & 41.4 & 50.7 & 64.5 & 71.1 & 50.2 & 52.3 & \cellcolor{gray!20}54.1 \\
    Qwen2.5-7B & 94.6 & 89.2 & 70.7 & 60.6 & 95.0 & 96.2 & 72.4 & \cellcolor{gray!20}82.7 \\
    Qwen2.5-14B & 81.7 & 69.5 & 69.2 & 69.4 & 90.8 & 63.0 & 72.6 & \cellcolor{gray!20}73.8 \\
    Qwen2.5-32B & 97.8 & 94.0 & 92.2 & 75.0 & 95.7 & 96.4 & 82.9 & \cellcolor{gray!20}90.6 \\
    Qwen2.5-72B & 98.8 & 95.8 & 92.4 & 76.4 & 96.9 & 98.0 & 88.7 & \cellcolor{gray!20}92.4 \\
    \bottomrule
  \end{tabular}
  \label{tab:full-res-jud_acc-math}
\end{table}

\begin{table}[htbp]
  \small
  \centering
  \caption{Full judge accuracy results of MMLU. The numbers in \Cref{fig:corre_task_acc_judge_acc} correspond to the average column.}
  \begin{tabular}{lrrrrrrr>{\columncolor{gray!20}}r}
    \toprule
    & \multicolumn{8}{c}{\textbf{Evaluatee}} \\
    \cmidrule{2-9}
    \textbf{Evaluator} & \multicolumn{1}{c}{\rotatebox{90}{Mistral-7b}} & \multicolumn{1}{c}{\rotatebox{90}{Phi-3.5}} & \multicolumn{1}{c}{\rotatebox{90}{GPT-3.5-Turbo}} & \multicolumn{1}{c}{\rotatebox{90}{GPT-4o}} & \multicolumn{1}{c}{\rotatebox{90}{Gemma-2-2B}} & \multicolumn{1}{c}{\rotatebox{90}{Llama-3.2-1B}} & \multicolumn{1}{c}{\rotatebox{90}{Mistral-Small}} & \multicolumn{1}{c}{\rotatebox{90}{Average}} \\
    \midrule
    Llama-3.2-3B & 58.1 & 37.7 & 53.3 & 60.0 & 48.2 & 43.6 & 71.5 & \cellcolor{gray!20}53.2 \\
    Llama-3.1-8B & 57.7 & 39.6 & 42.5 & 75.6 & 48.0 & 74.3 & 53.3 & \cellcolor{gray!20}55.9 \\
    Llama-3.1-70B & 73.3 & 53.7 & 63.0 & 78.9 & 72.0 & 84.4 & 65.6 & \cellcolor{gray!20}70.1 \\
    Llama-3.3-70B & 87.1 & 79.9 & 78.2 & 68.6 & 85.0 & 90.3 & 69.7 & \cellcolor{gray!20}79.8 \\
    \midrule
    Gemma-2-9B & 51.2 & 30.8 & 44.6 & 67.2 & 41.3 & 68.6 & 54.7 & \cellcolor{gray!20}51.2 \\
    Gemma-2-27B & 56.7 & 27.8 & 39.3 & 69.9 & 42.5 & 75.4 & 57.4 & \cellcolor{gray!20}52.7 \\
    \midrule
    Qwen2.5-3B & 45.9 & 37.9 & 45.7 & 41.9 & 59.1 & 52.4 & 48.4 & \cellcolor{gray!20}47.3 \\
    Qwen2.5-7B & 71.9 & 31.1 & 48.3 & 74.8 & 64.3 & 82.5 & 51.3 & \cellcolor{gray!20}60.6 \\
    Qwen2.5-14B & 83.0 & 70.8 & 67.7 & 64.6 & 73.7 & 85.0 & 61.8 & \cellcolor{gray!20}72.4 \\
    Qwen2.5-32B & 87.9 & 74.1 & 72.3 & 64.5 & 80.8 & 89.3 & 64.6 & \cellcolor{gray!20}76.2 \\
    Qwen2.5-72B & 90.8 & 72.8 & 70.6 & 60.4 & 85.4 & 90.8 & 75.7 & \cellcolor{gray!20}78.1 \\
    \bottomrule
  \end{tabular}
  \label{tab:full-res-jud_acc-mmlu}
\end{table}

\begin{table}[htbp]
  \small
  \centering
  \caption{Full judge accuracy results of MBPP+. The numbers in \Cref{fig:corre_task_acc_judge_acc} correspond to the average column.}
  \begin{tabular}{lrrrrrrr>{\columncolor{gray!20}}r}
    \toprule
    & \multicolumn{8}{c}{\textbf{Evaluatee}} \\
    \cmidrule{2-9}
    \textbf{Evaluator} & \multicolumn{1}{c}{\rotatebox{90}{Mistral-7b}} & \multicolumn{1}{c}{\rotatebox{90}{Phi-3.5}} & \multicolumn{1}{c}{\rotatebox{90}{GPT-3.5-Turbo}} & \multicolumn{1}{c}{\rotatebox{90}{GPT-4o}} & \multicolumn{1}{c}{\rotatebox{90}{Gemma-2-2B}} & \multicolumn{1}{c}{\rotatebox{90}{Llama-3.2-1B}} & \multicolumn{1}{c}{\rotatebox{90}{Mistral-Small}} & \multicolumn{1}{c}{\rotatebox{90}{Average}} \\
    \midrule
    Llama-3.2-3B & 37.3 & 51.9 & 60.2 & 48.9 & 60.0 & 46.8 & 54.1 & \cellcolor{gray!20}51.3 \\
    Llama-3.1-8B & 77.6 & 53.0 & 36.8 & 32.9 & 60.9 & 58.8 & 48.6 & \cellcolor{gray!20}52.6 \\
    Llama-3.1-70B & 80.0 & 62.8 & 45.2 & 47.3 & 75.0 & 85.8 & 58.8 & \cellcolor{gray!20}65.0 \\
    Llama-3.3-70B & 80.9 & 75.3 & 63.2 & 52.1 & 84.3 & 94.6 & 65.3 & \cellcolor{gray!20}73.7 \\
    \midrule
    Gemma-2-9B & 72.5 & 56.3 & 53.2 & 74.4 & 61.0 & 71.3 & 55.7 & \cellcolor{gray!20}63.5 \\
    Gemma-2-27B & 68.7 & 64.9 & 40.4 & 56.0 & 67.6 & 72.4 & 49.0 & \cellcolor{gray!20}59.9 \\
    \midrule
    Qwen2.5-3B & 47.5 & 54.4 & 55.1 & 44.3 & 49.5 & 47.1 & 47.5 & \cellcolor{gray!20}49.3 \\
    Qwen2.5-7B & 82.8 & 62.0 & 53.8 & 64.3 & 70.6 & 86.9 & 64.2 & \cellcolor{gray!20}69.2 \\
    Qwen2.5-14B & 88.0 & 74.3 & 60.9 & 53.7 & 77.4 & 88.5 & 69.4 & \cellcolor{gray!20}73.2 \\
    Qwen2.5-32B & 84.2 & 74.7 & 68.0 & 56.3 & 85.7 & 93.1 & 62.0 & \cellcolor{gray!20}74.8 \\
    Qwen2.5-72B & 91.3 & 82.3 & 69.1 & 42.1 & 86.7 & 93.9 & 71.4 & \cellcolor{gray!20}76.7 \\
    \bottomrule
  \end{tabular}
  \label{tab:full-res-jud_acc-mbpp}
\end{table}

\begin{table}[htbp]
  \small
  \centering
  \caption{Full LSPR results of MATH500. The numbers in \Cref{fig:corr_task_acc_legit_sp_ratio} correspond to the average column.}
  \begin{tabular}{lrrrrrrr>{\columncolor{gray!20}}r}
    \toprule
    & \multicolumn{8}{c}{\textbf{Evaluatee}} \\
    \cmidrule{2-9}
    \textbf{Evaluator} & \multicolumn{1}{c}{\rotatebox{90}{Mistral-7b}} & \multicolumn{1}{c}{\rotatebox{90}{Phi-3.5}} & \multicolumn{1}{c}{\rotatebox{90}{GPT-3.5-Turbo}} & \multicolumn{1}{c}{\rotatebox{90}{GPT-4o}} & \multicolumn{1}{c}{\rotatebox{90}{Gemma-2-2B}} & \multicolumn{1}{c}{\rotatebox{90}{Llama-3.2-1B}} & \multicolumn{1}{c}{\rotatebox{90}{Mistral-Small}} & \multicolumn{1}{c}{\rotatebox{90}{Average}} \\
    \midrule
    Llama-3.2-3B & 85.7 & 0.0 & 66.7 & 16.7 & 100.0 & 93.3 & 0.0 & \cellcolor{gray!20}51.8 \\
    Llama-3.1-8B & 94.0 & 90.0 & 83.3 & 0.0 & 90.4 & 95.0 & 80.0 & \cellcolor{gray!20}76.1 \\
    Llama-3.1-70B & 98.4 & 93.8 & 92.0 & 68.2 & 98.1 & 96.7 & 86.2 & \cellcolor{gray!20}90.5 \\
    Llama-3.3-70B & 98.5 & 97.5 & 97.5 & 93.3 & 99.4 & 98.2 & 91.4 & \cellcolor{gray!20}96.6 \\
    \midrule
    Gemma-2-9B & 98.5 & 62.5 & 79.2 & 16.7 & 98.3 & 91.1 & 50.0 & \cellcolor{gray!20}70.9 \\
    Gemma-2-27B & 97.2 & 84.1 & 91.7 & 37.5 & 95.6 & 89.0 & 73.3 & \cellcolor{gray!20}81.2 \\
    \midrule
    Qwen2.5-3B & 100.0 & 80.0 & 100.0 & 33.3 & 100.0 & 100.0 & 75.0 & \cellcolor{gray!20}84.0 \\
    Qwen2.5-7B & 99.5 & 95.9 & 96.5 & 66.7 & 99.4 & 98.9 & 97.6 & \cellcolor{gray!20}93.5 \\
    Qwen2.5-14B & 99.4 & 96.2 & 97.6 & 75.0 & 100.0 & 100.0 & 97.9 & \cellcolor{gray!20}95.2 \\
    Qwen2.5-32B & 99.0 & 99.3 & 98.1 & 92.3 & 99.2 & 98.8 & 97.1 & \cellcolor{gray!20}97.7 \\
    Qwen2.5-72B & 99.1 & 100.0 & 97.8 & 87.2 & 99.3 & 98.6 & 99.0 & \cellcolor{gray!20}97.3 \\
    \bottomrule
  \end{tabular}
  \label{tab:full-res-lspr-math}
\end{table}

\begin{table}[htbp]
  \centering
  \small
  \caption{Full LSPR results of MMLU. The numbers in \Cref{fig:corr_task_acc_legit_sp_ratio} correspond to the average column.}
  \begin{tabular}{lrrrrrrr>{\columncolor{gray!20}}r}
    \toprule
    & \multicolumn{8}{c}{\textbf{Evaluatee}} \\
    \cmidrule{2-9}
    \textbf{Evaluator} & \multicolumn{1}{c}{\rotatebox{90}{Mistral-7b}} & \multicolumn{1}{c}{\rotatebox{90}{Phi-3.5}} & \multicolumn{1}{c}{\rotatebox{90}{GPT-3.5-Turbo}} & \multicolumn{1}{c}{\rotatebox{90}{GPT-4o}} & \multicolumn{1}{c}{\rotatebox{90}{Gemma-2-2B}} & \multicolumn{1}{c}{\rotatebox{90}{Llama-3.2-1B}} & \multicolumn{1}{c}{\rotatebox{90}{Mistral-Small}} & \multicolumn{1}{c}{\rotatebox{90}{Average}} \\
    \midrule
    Llama-3.2-3B & 16.7 & 66.7 & 31.3 & 0.0 & 45.5 & 65.7 & 7.1 & \cellcolor{gray!20}33.3 \\
    Llama-3.1-8B & 71.8 & 84.6 & 53.0 & 16.7 & 81.5 & 83.4 & 37.5 & \cellcolor{gray!20}61.2 \\
    Llama-3.1-70B & 81.1 & 88.4 & 81.7 & 36.4 & 92.5 & 87.4 & 62.7 & \cellcolor{gray!20}75.7 \\
    Llama-3.3-70B & 92.3 & 97.2 & 91.3 & 68.8 & 96.9 & 93.7 & 81.3 & \cellcolor{gray!20}88.8 \\
    \midrule
    Gemma-2-9B & 83.3 & 88.1 & 75.0 & 18.2 & 90.6 & 84.7 & 30.8 & \cellcolor{gray!20}67.2 \\
    Gemma-2-27B & 84.4 & 84.2 & 68.5 & 9.1 & 95.9 & 87.7 & 53.6 & \cellcolor{gray!20}69.1 \\
    \midrule
    Qwen2.5-3B & 51.6 & 58.3 & 26.7 & 13.6 & 45.0 & 77.4 & 15.4 & \cellcolor{gray!20}41.2 \\
    Qwen2.5-7B & 84.4 & 75.9 & 68.7 & 33.3 & 86.1 & 86.9 & 50.6 & \cellcolor{gray!20}69.4 \\
    Qwen2.5-14B & 91.5 & 91.9 & 80.7 & 30.0 & 94.9 & 89.1 & 62.0 & \cellcolor{gray!20}77.2 \\
    Qwen2.5-32B & 90.9 & 95.5 & 82.9 & 35.3 & 93.8 & 90.6 & 69.6 & \cellcolor{gray!20}79.8 \\
    Qwen2.5-72B & 93.3 & 93.0 & 87.8 & 45.0 & 96.0 & 92.7 & 80.6 & \cellcolor{gray!20}84.1 \\
    \bottomrule
  \end{tabular}
  \label{tab:full-res-lspr-mmlu}
\end{table}

\begin{table}[htbp]
  \centering
  \small
  \caption{Full LSPR results of MBPP+. The numbers in \Cref{fig:corr_task_acc_legit_sp_ratio} correspond to the average column.}
  \begin{tabular}{lrrrrrrr>{\columncolor{gray!20}}r}
    \toprule
    & \multicolumn{8}{c}{\textbf{Evaluatee}} \\
    \cmidrule{2-9}
    \textbf{Evaluator} & \multicolumn{1}{c}{\rotatebox{90}{Mistral-7b}} & \multicolumn{1}{c}{\rotatebox{90}{Phi-3.5}} & \multicolumn{1}{c}{\rotatebox{90}{GPT-3.5-Turbo}} & \multicolumn{1}{c}{\rotatebox{90}{GPT-4o}} & \multicolumn{1}{c}{\rotatebox{90}{Gemma-2-2B}} & \multicolumn{1}{c}{\rotatebox{90}{Llama-3.2-1B}} & \multicolumn{1}{c}{\rotatebox{90}{Mistral-Small}} & \multicolumn{1}{c}{\rotatebox{90}{Average}} \\
    \midrule
    Llama-3.2-3B & 85.7 & 83.3 & 0.0 & 0.0 & 100.0 & 80.0 & 25.0 & \cellcolor{gray!20}53.4 \\
    Llama-3.1-8B & 100.0 & 12.5 & 50.0 & 0.0 & 85.7 & 80.0 & 11.1 & \cellcolor{gray!20}48.5 \\
    Llama-3.1-70B & 96.2 & 56.1 & 44.0 & 10.0 & 85.7 & 91.9 & 60.7 & \cellcolor{gray!20}63.5 \\
    Llama-3.3-70B & 97.5 & 86.7 & 73.7 & 30.8 & 91.8 & 100.0 & 86.5 & \cellcolor{gray!20}81.0 \\
    \midrule
    Gemma-2-9B & 96.6 & 37.5 & 29.4 & 14.3 & 93.3 & 95.5 & 50.0 & \cellcolor{gray!20}59.5 \\
    Gemma-2-27B & 100.0 & 100.0 & 44.4 & 66.7 & 96.3 & 100.0 & 100.0 & \cellcolor{gray!20}86.8 \\
    \midrule
    Qwen2.5-3B & 100.0 & 33.3 & 14.3 & 12.5 & 76.9 & 66.7 & 40.0 & \cellcolor{gray!20}49.1 \\
    Qwen2.5-7B & 96.7 & 82.6 & 63.2 & 25.0 & 91.1 & 96.3 & 85.7 & \cellcolor{gray!20}77.2 \\
    Qwen2.5-14B & 97.2 & 80.0 & 58.8 & 40.0 & 91.8 & 97.0 & 83.3 & \cellcolor{gray!20}78.3 \\
    Qwen2.5-32B & 97.2 & 90.0 & 78.3 & 75.0 & 93.2 & 97.3 & 96.3 & \cellcolor{gray!20}89.6 \\
    Qwen2.5-72B & 96.0 & 90.9 & 83.3 & 40.0 & 92.8 & 97.6 & 94.9 & \cellcolor{gray!20}85.1 \\
    \bottomrule
  \end{tabular}
  \label{tab:full-res-lspr-mbpp}
\end{table}

\begin{table}[htbp]
  \small
  \centering
  \caption{Full HSPP results of MATH500 at varying levels of evaluator reasoning. The numbers in \Cref{fig:reasoning_harmful_self-preference} correspond to the average column.}
  \begin{tabular}{lrrrrrrr>{\columncolor{gray!20}}r}
    \toprule
    & \multicolumn{8}{c}{\textbf{Evaluatee}} \\
    \cmidrule{2-9}
    \textbf{Evaluator} & \multicolumn{1}{c}{\rotatebox{90}{Mistral-7b}} & \multicolumn{1}{c}{\rotatebox{90}{Phi-3.5}} & \multicolumn{1}{c}{\rotatebox{90}{GPT-3.5-Turbo}} & \multicolumn{1}{c}{\rotatebox{90}{GPT-4o}} & \multicolumn{1}{c}{\rotatebox{90}{Gemma-2-2B}} & \multicolumn{1}{c}{\rotatebox{90}{Llama-3.2-1B}} & \multicolumn{1}{c}{\rotatebox{90}{Mistral-Small}} & \multicolumn{1}{c}{\rotatebox{90}{Average}} \\
    \midrule
    Llama-3.1-8B No reasoning & 43.5 & 32.1 & 22.8 & 19.0 & 43.5 & 29.2 & 16.7 & \cellcolor{gray!20}29.5 \\
    Llama-3.1-8B CoT & 39.1 & 19.6 & 26.3 & 6.5 & 30.4 & 25.0 & 12.2 & \cellcolor{gray!20}22.8 \\
    DeepSeek-R1-Distill-Llama-8B & 16.7 & 16.1 & 14.8 & 8.3 & 11.1 & 27.3 & 13.0 & \cellcolor{gray!20}15.3 \\
    \midrule
    Llama-3.3-70B No reasoning & 75.0 & 20.0 & 26.7 & 20.0 & 28.6 & 41.7 & 28.0 & \cellcolor{gray!20}34.3 \\
    Llama-3.3-70B CoT & 62.5 & 20.0 & 20.0 & 12.7 & 0.0 & 25.0 & 8.0 & \cellcolor{gray!20}21.2 \\
    DeepSeek-R1-Distill-Llama-70B & 50.0 & 0.0 & 44.4 & 0.0 & 0.0 & 0.0 & 28.6 & \cellcolor{gray!20}17.6 \\
    \midrule
    Qwen2.5-7B No reasoning & 66.7 & 72.7 & 27.3 & 51.3 & 80.0 & 62.5 & 38.1 & \cellcolor{gray!20}56.9 \\
    Qwen2.5-7B CoT & 33.3 & 9.1 & 45.5 & 10.3 & 20.0 & 12.5 & 4.8 & \cellcolor{gray!20}19.3 \\
    DeepSeek-R1-Distill-Qwen-7B & 66.7 & 7.1 & 7.1 & 9.4 & 0.0 & 14.3 & 14.3 & \cellcolor{gray!20}17.0 \\
    \midrule
    Qwen2.5-14B No reasoning & 80.0 & 83.3 & 71.4 & 23.3 & 100.0 & 16.7 & 41.7 & \cellcolor{gray!20}59.5 \\
    Qwen2.5-14B CoT & 60.0 & 16.7 & 57.1 & 13.3 & 0.0 & 33.3 & 8.3 & \cellcolor{gray!20}27.0 \\
    DeepSeek-R1-Distill-Qwen-14B & 60.0 & 7.7 & 25.0 & 11.5 & 16.7 & 25.0 & 22.2 & \cellcolor{gray!20}24.0 \\
    \midrule
    Qwen2.5-32B No reasoning & 100.0 & 100.0 & 80.0 & 33.3 & 75.0 & 71.4 & 37.5 & \cellcolor{gray!20}71.0 \\
    Qwen2.5-32B CoT & 100.0 & 50.0 & 100.0 & 11.1 & 50.0 & 57.1 & 25.0 & \cellcolor{gray!20}56.2 \\
    DeepSeek-R1-Distill-Qwen-32B & 100.0 & 40.0 & 50.0 & 0.0 & 0.0 & 25.0 & 27.3 & \cellcolor{gray!20}34.6 \\
    \bottomrule
  \end{tabular}
  \label{tab:full-res-hspp-math-reasoning}
\end{table}

\begin{table}[htbp]
  \small
  \centering
  \caption{Full HSPP results of MMLU at varying levels of evaluator reasoning. The numbers in \Cref{fig:reasoning_harmful_self-preference} correspond to the average column.}
  \begin{tabular}{lrrrrrrr>{\columncolor{gray!20}}r}
    \toprule
    & \multicolumn{8}{c}{\textbf{Evaluatee}} \\
    \cmidrule{2-9}
    \textbf{Evaluator} & \multicolumn{1}{c}{\rotatebox{90}{Mistral-7b}} & \multicolumn{1}{c}{\rotatebox{90}{Phi-3.5}} & \multicolumn{1}{c}{\rotatebox{90}{GPT-3.5-Turbo}} & \multicolumn{1}{c}{\rotatebox{90}{GPT-4o}} & \multicolumn{1}{c}{\rotatebox{90}{Gemma-2-2B}} & \multicolumn{1}{c}{\rotatebox{90}{Llama-3.2-1B}} & \multicolumn{1}{c}{\rotatebox{90}{Mistral-Small}} & \multicolumn{1}{c}{\rotatebox{90}{Average}} \\
    \midrule
    Llama-3.1-8B No reasoning & 66.3 & 39.3 & 54.3 & 12.0 & 34.5 & 88.0 & 42.6 & \cellcolor{gray!20}48.1 \\
    Llama-3.1-8B CoT & 60.5 & 26.8 & 50.5 & 2.8 & 24.1 & 86.7 & 37.2 & \cellcolor{gray!20}41.2 \\
    DeepSeek-R1-Distill-Llama-8B & 46.2 & 30.0 & 47.2 & 10.9 & 33.8 & 82.4 & 30.7 & \cellcolor{gray!20}40.2 \\
    \midrule
    Llama-3.3-70B No reasoning & 89.7 & 88.9 & 77.8 & 25.0 & 72.4 & 94.3 & 66.0 & \cellcolor{gray!20}73.4 \\
    Llama-3.3-70B CoT & 93.1 & 72.2 & 75.0 & 17.1 & 65.5 & 97.1 & 59.6 & \cellcolor{gray!20}68.5 \\
    DeepSeek-R1-Distill-Llama-70B & 70.8 & 52.6 & 71.0 & 24.5 & 70.0 & 86.2 & 58.1 & \cellcolor{gray!20}61.9 \\
    \midrule
    Qwen2.5-7B No reasoning & 81.4 & 55.2 & 58.4 & 13.5 & 75.0 & 98.3 & 63.7 & \cellcolor{gray!20}63.6 \\
    Qwen2.5-7B CoT & 54.2 & 24.1 & 29.9 & 4.1 & 39.6 & 96.7 & 34.5 & \cellcolor{gray!20}40.4 \\
    DeepSeek-R1-Distill-Qwen-7B & 39.8 & 31.4 & 32.9 & 9.4 & 25.3 & 63.5 & 29.5 & \cellcolor{gray!20}33.1 \\
    \midrule
    Qwen2.5-14B No reasoning & 68.4 & 70.0 & 80.0 & 31.3 & 66.7 & 98.1 & 64.6 & \cellcolor{gray!20}68.4 \\
    Qwen2.5-14B CoT & 76.3 & 60.0 & 78.0 & 12.2 & 63.3 & 98.1 & 55.4 & \cellcolor{gray!20}63.3 \\
    DeepSeek-R1-Distill-Qwen-14B & 72.7 & 37.9 & 59.3 & 8.0 & 63.3 & 85.1 & 45.3 & \cellcolor{gray!20}53.1 \\
    \midrule
    Qwen2.5-32B No reasoning & 89.5 & 77.8 & 94.0 & 27.2 & 73.3 & 98.0 & 72.4 & \cellcolor{gray!20}76.0 \\
    Qwen2.5-32B CoT & 78.9 & 50.0 & 64.0 & 12.0 & 53.3 & 90.0 & 51.7 & \cellcolor{gray!20}57.1 \\
    DeepSeek-R1-Distill-Qwen-32B & 63.0 & 50.0 & 73.5 & 18.5 & 52.2 & 81.3 & 52.6 & \cellcolor{gray!20}55.9 \\
    \bottomrule
  \end{tabular}
  \label{tab:full-res-hspp-mmlu-reasoning}
\end{table}

\begin{table}[htbp]
  \small
  \centering
  \caption{Full HSPP results of MBPP+ at varying levels of evaluator reasoning. The numbers in \Cref{fig:reasoning_harmful_self-preference} correspond to the average column.}
  \begin{tabular}{lrrrrrrr>{\columncolor{gray!20}}r}
    \toprule
    & \multicolumn{8}{c}{\textbf{Evaluatee}} \\
    \cmidrule{2-9}
    \textbf{Evaluator} & \multicolumn{1}{c}{\rotatebox{90}{Mistral-7B}} & \multicolumn{1}{c}{\rotatebox{90}{Phi-3.5}} & \multicolumn{1}{c}{\rotatebox{90}{GPT-3.5-Turbo}} & \multicolumn{1}{c}{\rotatebox{90}{GPT-4o}} & \multicolumn{1}{c}{\rotatebox{90}{Gemma-2-2B}} & \multicolumn{1}{c}{\rotatebox{90}{Llama-3.2-1B}} & \multicolumn{1}{c}{\rotatebox{90}{Mistral-Small}} & \multicolumn{1}{c}{\rotatebox{90}{Average}} \\
    \midrule
    Llama-3.1-8B No reasoning & 42.9 & 85.3 & 74.0 & 73.1 & 45.5 & 50.0 & 86.8 & \cellcolor{gray!20}65.4 \\
    Llama-3.1-8B CoT & 14.3 & 23.5 & 20.0 & 17.9 & 31.8 & 16.7 & 23.7 & \cellcolor{gray!20}21.1 \\
    DeepSeek-R1-Distill-Llama-8B & 8.3 & 8.9 & 15.5 & 11.6 & 9.7 & 22.2 & 14.3 & \cellcolor{gray!20}12.9 \\
    \midrule
    Llama-3.3-70B No reasoning & 37.5 & 53.3 & 45.5 & 41.4 & 64.3 & 16.7 & 50.0 & \cellcolor{gray!20}44.1 \\
    Llama-3.3-70B CoT & 37.5 & 40.0 & 22.7 & 27.6 & 71.4 & 16.7 & 27.8 & \cellcolor{gray!20}34.8 \\
    DeepSeek-R1-Distill-Llama-70B & 25.0 & 4.4 & 7.0 & 3.2 & 6.7 & 7.4 & 6.5 & \cellcolor{gray!20}8.6 \\
    \midrule
    Qwen2.5-7B No reasoning & 66.7 & 50.0 & 44.1 & 26.8 & 68.8 & 50.0 & 27.3 & \cellcolor{gray!20}47.7 \\
    Qwen2.5-7B CoT & 33.3 & 40.9 & 29.4 & 22.0 & 50.0 & 40.0 & 13.6 & \cellcolor{gray!20}32.7 \\
    DeepSeek-R1-Distill-Qwen-7B & 38.5 & 14.0 & 12.3 & 12.5 & 11.4 & 10.0 & 10.9 & \cellcolor{gray!20}15.7 \\
    \midrule
    Qwen2.5-14B No reasoning & 55.6 & 56.3 & 46.4 & 46.4 & 61.1 & 77.8 & 31.6 & \cellcolor{gray!20}53.6 \\
    Qwen2.5-14B CoT & 55.6 & 37.5 & 42.9 & 28.6 & 44.4 & 55.6 & 26.3 & \cellcolor{gray!20}41.5 \\
    DeepSeek-R1-Distill-Qwen-14B & 18.8 & 11.3 & 10.9 & 7.2 & 13.9 & 24.1 & 3.9 & \cellcolor{gray!20}12.9 \\
    \midrule
    Qwen2.5-32B No reasoning & 50.0 & 56.3 & 44.4 & 28.0 & 56.3 & 66.7 & 50.0 & \cellcolor{gray!20}50.2 \\
    Qwen2.5-32B CoT & 50.0 & 25.0 & 33.3 & 8.0 & 43.8 & 44.4 & 6.3 & \cellcolor{gray!20}30.1 \\
    DeepSeek-R1-Distill-Qwen-32B & 18.2 & 2.3 & 5.7 & 1.6 & 3.6 & 3.8 & 2.3 & \cellcolor{gray!20}5.4 \\
    \bottomrule
  \end{tabular}
  \label{tab:full-res-hspp-mbpp-reasoning}
\end{table}

\begin{table}[htbp]
  \small
  \centering
  \caption{Full HSPP token length results of MATH500. The numbers in \Cref{fig:reasoning_harmful_self-preference} correspond to the average column.}
  \begin{tabular}{lrrrrrrr>{\columncolor{gray!20}}r}
    \toprule
    & \multicolumn{8}{c}{\textbf{Evaluatee}} \\
    \cmidrule{2-9}
    \textbf{Evaluator} & \multicolumn{1}{c}{\rotatebox{90}{Mistral-7B}} & \multicolumn{1}{c}{\rotatebox{90}{Phi-3.5}} & \multicolumn{1}{c}{\rotatebox{90}{GPT-3.5-Turbo}} & \multicolumn{1}{c}{\rotatebox{90}{GPT-4o}} & \multicolumn{1}{c}{\rotatebox{90}{Gemma-2-2B}} & \multicolumn{1}{c}{\rotatebox{90}{Llama-3.2-1B}} & \multicolumn{1}{c}{\rotatebox{90}{Mistral-Small}} & \multicolumn{1}{c}{\rotatebox{90}{Average}} \\
    \midrule
    Llama-3.1-8B No reasoning & 0 & 0 & 0 & 0 & 0 & 0 & 0 & \cellcolor{gray!20}0 \\
    Llama-3.1-8B CoT & 175 & 195 & 191 & 171 & 176 & 156 & 161 & \cellcolor{gray!20}175 \\
    DeepSeek-R1-Distill-Llama-8B & 1545 & 1427 & 1596 & 1312 & 1431 & 1483 & 1517 & \cellcolor{gray!20}1473 \\
    \midrule
    Llama-3.3-70B No reasoning & 0 & 0 & 0 & 0 & 0 & 0 & 0 & \cellcolor{gray!20}0 \\
    Llama-3.3-70B CoT & 262 & 223 & 267 & 196 & 240 & 235 & 234 & \cellcolor{gray!20}237 \\
    DeepSeek-R1-Distill-Llama-70B & 790 & 842 & 948 & 798 & 771 & 815 & 912 & \cellcolor{gray!20}839 \\
    \midrule
    Qwen2.5-7B No reasoning & 0 & 0 & 0 & 0 & 0 & 0 & 0 & \cellcolor{gray!20}0 \\
    Qwen2.5-7B CoT & 103 & 95 & 105 & 85 & 101 & 94 & 95 & \cellcolor{gray!20}97 \\
    DeepSeek-R1-Distill-Qwen-7B & 821 & 828 & 878 & 784 & 779 & 890 & 893 & \cellcolor{gray!20}839 \\
    \midrule
    Qwen2.5-14B No reasoning & 0 & 0 & 0 & 0 & 0 & 0 & 0 & \cellcolor{gray!20}0 \\
    Qwen2.5-14B CoT & 153 & 142 & 158 & 131 & 155 & 151 & 146 & \cellcolor{gray!20}148 \\
    DeepSeek-R1-Distill-Qwen-14B & 904 & 921 & 1084 & 906 & 932 & 934 & 1028 & \cellcolor{gray!20}958 \\
    \midrule
    Qwen2.5-32B No reasoning & 0 & 0 & 0 & 0 & 0 & 0 & 0 & \cellcolor{gray!20}0 \\
    Qwen2.5-32B CoT & 131 & 125 & 132 & 111 & 134 & 128 & 123 & \cellcolor{gray!20}126 \\
    DeepSeek-R1-Distill-Qwen-32B & 795 & 768 & 891 & 731 & 728 & 759 & 904 & \cellcolor{gray!20}796 \\
    \bottomrule
  \end{tabular}
  \label{tab:full-res-hspp-math-reasoning-token}
\end{table}

\begin{table}[htbp]
  \small
  \centering
  \caption{Full HSPP token length results of MMLU. The numbers in \Cref{fig:reasoning_harmful_self-preference} correspond to the average column.}
  \begin{tabular}{lrrrrrrr>{\columncolor{gray!20}}r}
    \toprule
    & \multicolumn{8}{c}{\textbf{Evaluatee}} \\
    \cmidrule{2-9}
    \textbf{Evaluator} & \multicolumn{1}{c}{\rotatebox{90}{Mistral-7B}} & \multicolumn{1}{c}{\rotatebox{90}{Phi-3.5}} & \multicolumn{1}{c}{\rotatebox{90}{GPT-3.5-Turbo}} & \multicolumn{1}{c}{\rotatebox{90}{GPT-4o}} & \multicolumn{1}{c}{\rotatebox{90}{Gemma-2-2B}} & \multicolumn{1}{c}{\rotatebox{90}{Llama-3.2-1B}} & \multicolumn{1}{c}{\rotatebox{90}{Mistral-Small}} & \multicolumn{1}{c}{\rotatebox{90}{Average}} \\
    \midrule
    Llama-3.1-8B No reasoning & 0 & 0 & 0 & 0 & 0 & 0 & 0 & \cellcolor{gray!20}0 \\
    Llama-3.1-8B CoT & 172 & 183 & 171 & 183 & 187 & 171 & 173 & \cellcolor{gray!20}177 \\
    DeepSeek-R1-Distill-Llama-8B & 509 & 523 & 533 & 478 & 510 & 503 & 516 & \cellcolor{gray!20}510 \\
    \midrule
    Llama-3.3-70B No reasoning & 0 & 0 & 0 & 0 & 0 & 0 & 0 & \cellcolor{gray!20}0 \\
    Llama-3.3-70B CoT & 286 & 303 & 292 & 319 & 304 & 278 & 291 & \cellcolor{gray!20}296 \\
    DeepSeek-R1-Distill-Llama-70B & 410 & 419 & 417 & 384 & 431 & 400 & 407 & \cellcolor{gray!20}410 \\
    \midrule
    Qwen2.5-7B No reasoning & 0 & 0 & 0 & 0 & 0 & 0 & 0 & \cellcolor{gray!20}0 \\
    Qwen2.5-7B CoT & 108 & 111 & 112 & 108 & 108 & 103 & 110 & \cellcolor{gray!20}108 \\
    DeepSeek-R1-Distill-Qwen-7B & 467 & 458 & 468 & 441 & 513 & 488 & 453 & \cellcolor{gray!20}470 \\
    \midrule
    Qwen2.5-14B No reasoning & 0 & 0 & 0 & 0 & 0 & 0 & 0 & \cellcolor{gray!20}0 \\
    Qwen2.5-14B CoT & 134 & 138 & 134 & 141 & 144 & 136 & 134 & \cellcolor{gray!20}137 \\
    DeepSeek-R1-Distill-Qwen-14B & 518 & 516 & 517 & 464 & 525 & 528 & 497 & \cellcolor{gray!20}509 \\
    \midrule
    Qwen2.5-32B No reasoning & 0 & 0 & 0 & 0 & 0 & 0 & 0 & \cellcolor{gray!20}0 \\
    Qwen2.5-32B CoT & 126 & 127 & 125 & 131 & 135 & 132 & 136 & \cellcolor{gray!20}130 \\
    DeepSeek-R1-Distill-Qwen-32B & 501 & 512 & 501 & 475 & 512 & 489 & 487 & \cellcolor{gray!20}496 \\
    \bottomrule
  \end{tabular}
  \label{tab:full-res-hspp-mmlu-reasoning-token}
\end{table}

\begin{table}[htbp]
  \small
  \centering
  \caption{Full HSPP token length results of MBPP+. The numbers in \Cref{fig:reasoning_harmful_self-preference} correspond to the average column.}
  \begin{tabular}{lrrrrrrr>{\columncolor{gray!20}}r}
    \toprule
    & \multicolumn{8}{c}{\textbf{Evaluatee}} \\
    \cmidrule{2-9}
    \textbf{Evaluator} & \multicolumn{1}{c}{\rotatebox{90}{Mistral-7B}} & \multicolumn{1}{c}{\rotatebox{90}{Phi-3.5}} & \multicolumn{1}{c}{\rotatebox{90}{GPT-3.5-Turbo}} & \multicolumn{1}{c}{\rotatebox{90}{GPT-4o}} & \multicolumn{1}{c}{\rotatebox{90}{Gemma-2-2B}} & \multicolumn{1}{c}{\rotatebox{90}{Llama-3.2-1B}} & \multicolumn{1}{c}{\rotatebox{90}{Mistral-Small}} & \multicolumn{1}{c}{\rotatebox{90}{Average}} \\
    \midrule
    Llama-3.1-8B No reasoning & 0 & 0 & 0 & 0 & 0 & 0 & 0 & \cellcolor{gray!20}0 \\
    Llama-3.1-8B CoT & 428 & 388 & 382 & 343 & 396 & 344 & 371 & \cellcolor{gray!20}379 \\
    DeepSeek-R1-Distill-Llama-8B & 1435 & 1310 & 1336 & 1168 & 1429 & 1312 & 1305 & \cellcolor{gray!20}1328 \\
    \midrule
    Llama-3.3-70B No reasoning & 0 & 0 & 0 & 0 & 0 & 0 & 0 & \cellcolor{gray!20}0 \\
    Llama-3.3-70B CoT & 265 & 231 & 230 & 229 & 250 & 240 & 226 & \cellcolor{gray!20}239 \\
    DeepSeek-R1-Distill-Llama-70B & 1340 & 1215 & 1169 & 1097 & 1286 & 1247 & 1200 & \cellcolor{gray!20}1222 \\
    \midrule
    Qwen2.5-7B No reasoning & 0 & 0 & 0 & 0 & 0 & 0 & 0 & \cellcolor{gray!20}0 \\
    Qwen2.5-7B CoT & 105 & 102 & 101 & 101 & 109 & 101 & 99 & \cellcolor{gray!20}103 \\
    DeepSeek-R1-Distill-Qwen-7B & 1308 & 1314 & 1351 & 1249 & 1408 & 1273 & 1346 & \cellcolor{gray!20}1321 \\
    \midrule
    Qwen2.5-14B No reasoning & 0 & 0 & 0 & 0 & 0 & 0 & 0 & \cellcolor{gray!20}0 \\
    Qwen2.5-14B CoT & 161 & 157 & 169 & 162 & 160 & 155 & 168 & \cellcolor{gray!20}162 \\
    DeepSeek-R1-Distill-Qwen-14B & 1458 & 1328 & 1296 & 1244 & 1427 & 1402 & 1308 & \cellcolor{gray!20}1352 \\
    \midrule
    Qwen2.5-32B No reasoning & 0 & 0 & 0 & 0 & 0 & 0 & 0 & \cellcolor{gray!20}0 \\
    Qwen2.5-32B CoT & 599 & 524 & 556 & 403 & 578 & 513 & 466 & \cellcolor{gray!20}520 \\
    DeepSeek-R1-Distill-Qwen-32B & 1677 & 1567 & 1440 & 1269 & 1624 & 1523 & 1510 & \cellcolor{gray!20}1516 \\
    \bottomrule
  \end{tabular}
  \label{tab:full-res-hspp-mbpp-reasoning-token}
\end{table}

\begin{table}[htbp]
  \small
  \centering
  \caption{\textcolor{black}{Full SPR results of MATH500 under $\mathcal{D}_{\text{diff}}$. The numbers in \Cref{fig:d_diff_vs_d_agree} correspond to the average column.}}
  \begin{tabular}{
    >{\color{black}}l
    >{\color{black}}r
    >{\color{black}}r
    >{\color{black}}r
    >{\color{black}}r
    >{\color{black}}r
    >{\color{black}}r
    >{\color{black}}r
    >{\columncolor{gray!20}}>{\color{black}}r
  }
    \toprule
    & \multicolumn{8}{c}{\textbf{Evaluatee}} \\
    \cmidrule{2-9}
    \textbf{Evaluator} &
    \rotatebox{90}{Mistral-7b} &
    \rotatebox{90}{Phi-3.5} &
    \rotatebox{90}{GPT-3.5-Turbo} &
    \rotatebox{90}{GPT-4o} &
    \rotatebox{90}{Gemma-2-2B} &
    \rotatebox{90}{Llama-3.2-1B} &
    \rotatebox{90}{Mistral-Small} &
    \rotatebox{90}{Average} \\
    \midrule
    Llama-3.2-3B   & 4.27 & 3.70 & 2.21 & 6.03 & 5.85 & 21.28 & 2.04 & 6.48 \\
    Llama-3.1-8B   & 26.18 & 16.53 & 12.95 & 4.07 & 27.37 & 12.05 & 11.19 & 15.76 \\
    Llama-3.1-70B  & 80.00 & 58.82 & 50.00 & 18.80 & 66.80 & 69.23 & 26.13 & 52.83 \\
    Llama-3.3-70B  & 74.18 & 56.64 & 47.34 & 16.67 & 64.34 & 66.02 & 29.66 & 50.69 \\
    \midrule
    Gemma-2-9B     & 36.61 & 14.95 & 20.17 & 3.85 & 34.71 & 26.47 & 10.34 & 21.01 \\
    Gemma-2-27B    & 52.74 & 38.26 & 35.56 & 11.27 & 45.92 & 51.55 & 27.27 & 37.51 \\
    \midrule
    Qwen2.5-3B     & 2.92 & 3.91 & 4.11 & 2.80 & 2.09 & 1.81 & 3.67 & 3.04 \\
    Qwen2.5-7B     & 77.70 & 61.39 & 49.43 & 16.90 & 62.63 & 70.88 & 33.33 & 53.18 \\
    Qwen2.5-14B    & 58.80 & 47.31 & 44.86 & 27.78 & 61.22 & 35.87 & 37.90 & 44.82 \\
    Qwen2.5-32B    & 95.33 & 76.50 & 78.54 & 36.11 & 80.12 & 85.10 & 50.00 & 71.67 \\
    Qwen2.5-72B    & 97.86 & 88.89 & 86.26 & 65.28 & 82.52 & 91.78 & 69.72 & 83.18 \\
    \bottomrule
  \end{tabular}
  \label{tab:full-res-spr-math_d_diff}
\end{table}

\begin{table}[htbp]
  \small
  \centering
  \caption{\textcolor{black}{Full SPR results of MATH500 under $\mathcal{D}_{\text{agree}}$. The numbers in \Cref{fig:d_diff_vs_d_agree} correspond to the average column.}}
  \begin{tabular}{
    >{\color{black}}l
    >{\color{black}}r
    >{\color{black}}r
    >{\color{black}}r
    >{\color{black}}r
    >{\color{black}}r
    >{\color{black}}r
    >{\color{black}}r
    >{\columncolor{gray!20}}>{\color{black}}r
  }
    \toprule
    & \multicolumn{8}{c}{\textbf{Evaluatee}} \\
    \cmidrule{2-9}
    \textbf{Evaluator} &
    \rotatebox{90}{Mistral-7b} &
    \rotatebox{90}{Phi-3.5} &
    \rotatebox{90}{GPT-3.5-Turbo} &
    \rotatebox{90}{GPT-4o} &
    \rotatebox{90}{Gemma-2-2B} &
    \rotatebox{90}{Llama-3.2-1B} &
    \rotatebox{90}{Mistral-Small} &
    \rotatebox{90}{Average} \\
    \midrule
    Llama-3.2-3B   & 4.17 & 6.38 & 0.55 & 15.95 & 13.68 & 15.60 & 4.53 & 8.69 \\
    Llama-3.1-8B   & 17.15 & 11.61 & 4.99 & 5.79 & 20.00 & 11.38 & 4.92 & 10.83 \\
    Llama-3.1-70B  & 57.69 & 28.02 & 28.00 & 10.70 & 51.74 & 46.24 & 14.14 & 33.79 \\
    Llama-3.3-70B  & 56.89 & 25.77 & 23.87 & 7.56 & 47.81 & 35.25 & 14.40 & 30.22 \\
    \midrule
    Gemma-2-9B     & 18.61 & 7.38 & 8.66 & 7.27 & 24.85 & 16.36 & 3.65 & 12.40 \\
    Gemma-2-27B    & 30.77 & 15.58 & 14.79 & 6.98 & 45.07 & 30.39 & 8.72 & 21.76 \\
    \midrule
    Qwen2.5-3B     & 0.38 & 0.27 & 1.69 & 0.76 & 1.92 & 0.72 & 1.79 & 1.08 \\
    Qwen2.5-7B     & 40.99 & 19.01 & 14.11 & 5.13 & 43.38 & 35.15 & 6.90 & 23.52 \\
    Qwen2.5-14B    & 45.23 & 17.72 & 19.05 & 5.14 & 47.57 & 30.80 & 9.04 & 24.94 \\
    Qwen2.5-32B    & 59.78 & 24.92 & 32.54 & 7.71 & 53.93 & 50.00 & 18.33 & 35.32 \\
    Qwen2.5-72B    & 64.74 & 27.33 & 34.60 & 7.48 & 54.02 & 44.39 & 15.92 & 35.50 \\
    \bottomrule
  \end{tabular}
  \label{tab:full-res-spr-math-d_agree}
\end{table}

\begin{table}[htbp]
  \small
  \centering
  \caption{\textcolor{black}{Full SPR results of MMLU under $\mathcal{D}_{\text{diff}}$. The numbers in \Cref{fig:d_diff_vs_d_agree} correspond to the average column.}}
  \begin{tabular}{
    >{\color{black}}l
    >{\color{black}}r
    >{\color{black}}r
    >{\color{black}}r
    >{\color{black}}r
    >{\color{black}}r
    >{\color{black}}r
    >{\color{black}}r
    >{\columncolor{gray!20}}>{\color{black}}r
  }
    \toprule
    & \multicolumn{8}{c}{\textbf{Evaluatee}} \\
    \cmidrule{2-9}
    \textbf{Evaluator} &
    \rotatebox{90}{Mistral-7B} &
    \rotatebox{90}{Phi-3.5} &
    \rotatebox{90}{GPT-3.5-Turbo} &
    \rotatebox{90}{GPT-4o} &
    \rotatebox{90}{Gemma-2-2B} &
    \rotatebox{90}{Llama-3.2-1B} &
    \rotatebox{90}{Mistral-Small} &
    \rotatebox{90}{Average} \\
    \midrule
    Llama-3.2-3B   & 1.81 & 4.19 & 4.02 & 6.40 & 3.10 & 9.78 & 3.22 & 4.64 \\
    Llama-3.1-8B   & 38.27 & 11.06 & 18.23 & 2.29 & 17.06 & 50.78 & 17.47 & 22.17 \\
    Llama-3.1-70B  & 69.21 & 24.67 & 42.86 & 5.95 & 49.54 & 81.28 & 38.07 & 44.51 \\
    Llama-3.3-70B  & 66.95 & 52.82 & 57.18 & 13.56 & 56.83 & 80.81 & 55.15 & 54.76 \\
    \midrule
    Gemma-2-9B     & 36.36 & 8.99 & 16.94 & 6.21 & 17.49 & 52.39 & 15.29 & 21.95 \\
    Gemma-2-27B    & 36.58 & 8.12 & 17.53 & 6.25 & 20.00 & 60.96 & 16.57 & 23.72 \\
    \midrule
    Qwen2.5-3B     & 10.95 & 5.73 & 9.58 & 6.27 & 6.49 & 7.06 & 8.55 & 7.81 \\
    Qwen2.5-7B     & 61.94 & 11.84 & 23.99 & 7.14 & 41.40 & 75.00 & 40.70 & 37.43 \\
    Qwen2.5-14B    & 63.09 & 25.46 & 34.90 & 13.61 & 42.75 & 68.17 & 51.97 & 42.85 \\
    Qwen2.5-32B    & 81.18 & 57.11 & 59.34 & 13.71 & 58.81 & 85.49 & 63.35 & 59.86 \\
    Qwen2.5-72B    & 86.46 & 37.15 & 55.86 & 21.98 & 54.13 & 91.17 & 66.22 & 58.99 \\
    \bottomrule
  \end{tabular}
  \label{tab:full-res-spr-mmlu_d_diff}
\end{table}

\begin{table}[htbp]
  \small
  \centering
  \caption{\textcolor{black}{Full SPR results of MMLU under $\mathcal{D}_{\text{agree}}$. The numbers in \Cref{fig:d_diff_vs_d_agree} correspond to the average column.}}
  \begin{tabular}{
    >{\color{black}}l
    >{\color{black}}r
    >{\color{black}}r
    >{\color{black}}r
    >{\color{black}}r
    >{\color{black}}r
    >{\color{black}}r
    >{\color{black}}r
    >{\columncolor{gray!20}}>{\color{black}}r
  }
    \toprule
    & \multicolumn{8}{c}{\textbf{Evaluatee}} \\
    \cmidrule{2-9}
    \textbf{Evaluator} &
    \rotatebox{90}{Mistral-7B} &
    \rotatebox{90}{Phi-3.5} &
    \rotatebox{90}{GPT-3.5-Turbo} &
    \rotatebox{90}{GPT-4o} &
    \rotatebox{90}{Gemma-2-2B} &
    \rotatebox{90}{Llama-3.2-1B} &
    \rotatebox{90}{Mistral-Small} &
    \rotatebox{90}{Average} \\
    \midrule
    Llama-3.2-3B   & 2.40 & 4.39 & 4.32 & 10.20 & 5.27 & 7.01 & 3.89 & 5.35 \\
    Llama-3.1-8B   & 36.69 & 12.64 & 20.53 & 2.71 & 18.90 & 54.90 & 17.64 & 23.43 \\
    Llama-3.1-70B  & 60.19 & 19.92 & 34.53 & 2.09 & 31.87 & 79.78 & 24.04 & 36.06 \\
    Llama-3.3-70B  & 78.83 & 48.50 & 69.12 & 12.81 & 57.51 & 82.41 & 61.32 & 58.64 \\
    \midrule
    Gemma-2-9B     & 21.76 & 8.44 & 12.55 & 1.09 & 9.15 & 35.01 & 9.40 & 13.92 \\
    Gemma-2-27B    & 19.23 & 9.40 & 15.46 & 5.83 & 9.76 & 36.04 & 10.71 & 15.20 \\
    \midrule
    Qwen2.5-3B     & 5.30 & 5.51 & 4.37 & 0.92 & 3.90 & 5.17 & 4.17 & 4.19 \\
    Qwen2.5-7B     & 40.14 & 11.40 & 11.47 & 1.39 & 18.86 & 78.96 & 16.98 & 25.60 \\
    Qwen2.5-14B    & 63.84 & 23.00 & 41.12 & 3.63 & 34.81 & 85.46 & 31.49 & 40.48 \\
    Qwen2.5-32B    & 86.82 & 58.50 & 71.23 & 8.56 & 63.77 & 92.29 & 67.34 & 64.07 \\
    Qwen2.5-72B    & 76.26 & 33.33 & 56.24 & 2.53 & 43.15 & 95.38 & 48.47 & 50.77 \\
    \bottomrule
  \end{tabular}
  \label{tab:full-res-spr-mmlu_d_agree}
\end{table}

\begin{table}[htbp]
  \small
  \centering
  \caption{\textcolor{black}{Full SPR results of MBPP+ under $\mathcal{D}_{\text{diff}}$. The numbers in \Cref{fig:d_diff_vs_d_agree} correspond to the average column.}}
  \begin{tabular}{
    >{\color{black}}l
    >{\color{black}}r
    >{\color{black}}r
    >{\color{black}}r
    >{\color{black}}r
    >{\color{black}}r
    >{\color{black}}r
    >{\color{black}}r
    >{\columncolor{gray!20}}>{\color{black}}r
  }
    \toprule
    & \multicolumn{8}{c}{\textbf{Evaluatee}} \\
    \cmidrule{2-9}
    \textbf{Evaluator} &
    \rotatebox{90}{Mistral-7B} &
    \rotatebox{90}{Phi-3.5} &
    \rotatebox{90}{GPT-3.5-Turbo} &
    \rotatebox{90}{GPT-4o} &
    \rotatebox{90}{Gemma-2-2B} &
    \rotatebox{90}{Llama-3.2-1B} &
    \rotatebox{90}{Mistral-Small} &
    \rotatebox{90}{Average} \\
    \midrule
    Llama-3.2-3B   & 4.93 & 7.41 & 2.15 & 5.56 & 2.11 & 10.64 & 4.71 & 5.36 \\
    Llama-3.1-8B   & 9.52 & 12.12 & 11.76 & 11.39 & 7.61 & 10.31 & 12.86 & 10.80 \\
    Llama-3.1-70B  & 59.43 & 52.56 & 40.32 & 36.36 & 78.45 & 77.95 & 41.18 & 55.18 \\
    Llama-3.3-70B  & 61.86 & 61.64 & 33.33 & 27.08 & 80.17 & 83.85 & 49.33 & 56.75 \\
    \midrule
    Gemma-2-9B     & 18.13 & 22.54 & 22.08 & 8.54 & 19.48 & 20.37 & 19.67 & 18.69 \\
    Gemma-2-27B    & 23.08 & 24.56 & 15.79 & 6.00 & 25.71 & 31.90 & 9.80 & 19.55 \\
    \midrule
    Qwen2.5-3B     & 12.03 & 5.26 & 10.14 & 11.43 & 12.38 & 2.88 & 8.20 & 8.90 \\
    Qwen2.5-7B     & 50.56 & 32.39 & 29.23 & 14.29 & 51.38 & 67.21 & 31.34 & 39.49 \\
    Qwen2.5-14B    & 55.50 & 42.86 & 26.56 & 12.20 & 68.55 & 75.57 & 41.67 & 46.13 \\
    Qwen2.5-32B    & 52.97 & 48.19 & 30.67 & 8.33 & 66.17 & 78.47 & 34.18 & 45.57 \\
    Qwen2.5-72B    & 60.58 & 55.70 & 32.73 & 26.32 & 82.22 & 83.78 & 50.65 & 56.00 \\
    \bottomrule
  \end{tabular}
  \label{tab:full-res-spr-mbpp_d_diff}
\end{table}

\begin{table}[htbp]
  \small
  \centering
  \caption{\textcolor{black}{Full SPR results of MBPP+ under $\mathcal{D}_{\text{agree}}$. The numbers in \Cref{fig:d_diff_vs_d_agree} correspond to the average column.}}
  \begin{tabular}{
    >{\color{black}}l
    >{\color{black}}r
    >{\color{black}}r
    >{\color{black}}r
    >{\color{black}}r
    >{\color{black}}r
    >{\color{black}}r
    >{\color{black}}r
    >{\columncolor{gray!20}}>{\color{black}}r
  }
    \toprule
    & \multicolumn{8}{c}{\textbf{Evaluatee}} \\
    \cmidrule{2-9}
    \textbf{Evaluator} &
    \rotatebox{90}{Mistral-7B} &
    \rotatebox{90}{Phi-3.5} &
    \rotatebox{90}{GPT-3.5-Turbo} &
    \rotatebox{90}{GPT-4o} &
    \rotatebox{90}{Gemma-2-2B} &
    \rotatebox{90}{Llama-3.2-1B} &
    \rotatebox{90}{Mistral-Small} &
    \rotatebox{90}{Average} \\
    \midrule
    Llama-3.2-3B   & 7.20 & 8.42 & 4.21 & 6.60 & 7.07 & 7.04 & 8.19 & 6.96 \\
    Llama-3.1-8B   & 10.82 & 7.05 & 4.84 & 6.69 & 5.59 & 4.63 & 8.77 & 6.91 \\
    Llama-3.1-70B  & 38.92 & 25.33 & 25.00 & 16.10 & 40.84 & 44.22 & 21.94 & 30.34 \\
    Llama-3.3-70B  & 42.39 & 17.70 & 18.38 & 13.33 & 36.19 & 42.74 & 17.82 & 26.94 \\
    \midrule
    Gemma-2-9B     & 15.14 & 11.07 & 9.63 & 8.11 & 12.96 & 14.81 & 6.62 & 11.19 \\
    Gemma-2-27B    & 13.27 & 7.79 & 7.17 & 4.57 & 9.52 & 11.07 & 4.59 & 8.28 \\
    \midrule
    Qwen2.5-3B     & 10.00 & 6.23 & 13.27 & 8.77 & 15.75 & 13.50 & 10.41 & 11.13 \\
    Qwen2.5-7B     & 31.82 & 25.41 & 19.49 & 13.66 & 27.88 & 30.08 & 21.22 & 24.22 \\
    Qwen2.5-14B    & 32.09 & 14.94 & 12.74 & 5.64 & 27.95 & 31.17 & 10.46 & 19.28 \\
    Qwen2.5-32B    & 26.14 & 9.49 & 5.28 & 1.21 & 23.67 & 25.21 & 9.36 & 14.34 \\
    Qwen2.5-72B    & 35.29 & 15.72 & 9.29 & 4.71 & 30.04 & 33.04 & 12.29 & 20.05 \\
    \bottomrule
  \end{tabular}
  \label{tab:full-res-spr-mbpp_d_agree}
\end{table}

\clearpage

\color{black}{
\section{Limitations}
\label{sec:limit}
\paragraph{Biases and hacking other than self-preference.}
While we focus on self-preference bias, LLM evaluators may also exhibit other forms of bias, such as length bias~\citep{saito2023verbosity,zheng2023judging}, order bias~\citep{zheng2023large,wang2023large-e6b,pezeshkpour2024large,wei2024unveiling,shi2024judging}, or stylistic bias~\citep{chen2024humans,ye2024justice}, which undermine the fairness of LLM-based evaluations.
In addition,~\citet{zheng2025cheating} uncover potential hacking in LLM-as-a-Judge benchmarking, showing that simple ``null" models that always produce constant outputs can cheat automatic evaluation and achieve top-performing win rates.
Collectively, these findings highlight the need for more comprehensive audits to ensure robustness and reliability for LLM-based evaluations.
\paragraph{Diverse inference-time scaling strategies.}
Our experiments on varying levels of CoT reasoning length represent a subset of inference-time scaling methods~\citep{welleck2024from-159}.
Future work can explore a broader space of techniques for scaling evaluation-time compute, such as self-consistency decoding~\citep{wang2022self-consistency-e4c}, Best-of-N via repeated sampling~\citep{brown2024large-581,chen2024are-636}, and multi-agent verification~\citep{lifshitz2025multi}.
A notable concurrent work by~\citet{kim2025scaling} explores using reasoning models as process evaluators and demonstrates its superiority over process reward models.
In comparison, we focus on investigating self-evaluation in setups analogous to LLM-as-a-Judge.
\paragraph{Broader LLM-based evaluation formats.}
While our experiments investigate self-preference strictly as a model favoring its exact own output, a more relaxed definition may include preference toward outputs from models within the same family.
Exploring this broader notion could uncover subtler forms of bias that arise from shared training signals or architectural similarities.
To enable consistent measurement of self-preference behaviors, we adopt a controlled setup that separates evaluators and evaluatees and focuses on discrete, pairwise evaluation.
However, applications such as scalar scoring or ranking might involve pointwise evaluation~\citep{lee2023rlaif,foundation-autoraters,zhang2024generative}.
Extending our methodology to these alternative formats would help generalize our findings and offer a more comprehensive understanding of self-preference in LLM-based evaluations.
}

\end{document}